\pdfoutput=1

\documentclass[11pt]{article}

\usepackage{EMNLP2023}  

\usepackage{times}
\usepackage{latexsym}

\usepackage[T1]{fontenc}

\usepackage[utf8]{inputenc}

\usepackage{microtype}

\usepackage{inconsolata}

\usepackage{algorithm}
\usepackage[noend]{algpseudocode}
\usepackage{amsfonts} 
\usepackage{amsmath}
\usepackage{amssymb}
\usepackage{bbm}
\usepackage{booktabs}
\usepackage{float}
\usepackage{graphicx}
\usepackage{hyperref}
\usepackage{makecell}
\usepackage{multirow}
\usepackage{subcaption}
\usepackage{threeparttable} 
\usepackage{url}
\usepackage{xcolor}

\title{Boosting Summarization with Normalizing Flows and Aggressive Training}

\author{Yu Yang \\
  University of Minnesota \\
  \texttt{yang6367@umn.edu} \\\And
  Xiaotong Shen \\
  University of Minnesota \\
  \texttt{xshen@umn.edu} \\}

\begin{document}
\maketitle
\begin{abstract}
This paper presents FlowSUM, a normalizing flows-based variational encoder-decoder framework for Transformer-based summarization. Our approach tackles two primary challenges in variational summarization: insufficient semantic information in latent representations and posterior collapse during training. To address these challenges, we employ normalizing flows to enable flexible latent posterior modeling, and we propose a controlled alternate aggressive training (CAAT) strategy with an improved gate mechanism. Experimental results show that FlowSUM significantly enhances the quality of generated summaries and unleashes the potential for knowledge distillation with minimal impact on inference time. Furthermore, we investigate the issue of posterior collapse in normalizing flows and analyze how the summary quality is affected by the training strategy, gate initialization, and the type and number of normalizing flows used, offering valuable insights for future research.
\end{abstract}

\section{Introduction}\label{sec:intro}
Abstractive summarization \citep{see-etal-2017-get, paulus2018deep, wang2018reinforced} aims to generate summaries by rephrasing or introducing novel words to capture the most salient information in the source text. Many abstractive summarization models \citep{liu-lapata-2019-text, zhang2020pegasus, rothe-etal-2020-leveraging, raffel2020exploring} are based on the Transformers architecture \citep{vaswani2017attention} and have consistently produced state-of-the-art summarization quality. However, issues such as exposure bias \citep{ranzato2016sequence, qi-etal-2020-prophetnet}, lack of text generation diversity \citep{holtzman2020curious}, and insufficient capturing of semantic information \citep{reimers-gurevych-2019-sentence, wang-etal-2020-friendly} remain.

Variational models have gained increasing research interest \citep{zhang-etal-2016-variational-neural, su2018variational, wang-etal-2019-topic, fu2020document} as they address these issues by introducing uncertainty in predictions through learning a probability distribution over latent variables. A variational model enables diverse text generation \citep{du2022diverse}, smoother output spaces, and semantically meaningful latent codes \citep{wang-etal-2019-topic} that guide the generation of coherent and informative summaries.

Nonetheless, existing variational models have not fully achieved the aforementioned desirable properties due to two main challenges. Firstly, the semantic information in the source text may possess a complex structure. However, since introducing latent variables complicates parameter estimation, many current models \citep{fu2020document, zheng2020topic} represent latent codes using a Gaussian distribution, which is insufficient for capturing the intricacies of the latent space and could potentially reduce model performance. To enrich latent distributions, researchers suggest replacing the highly restricted isotropic Gaussian with normalizing flows \citep{rezende2015variational}. Normalizing flows can generate complex distributions while preserving density in an analytical form, and they have been integrated into variational autoencoder (VAE) \citep{kingma2014auto, rezende2014stochastic} and variational encoder-decoder (VED) \citep{serban2017hierarchical, zhou-neubig-2017-morphological} frameworks to better approximate the latent posterior. This approach has found application in various domains, including text generation \citep{wang-etal-2019-topic}, neural machine translation \citep{setiawan-etal-2020-variational}, and dialogue generation \citep{luo2021variational}. Despite this progress, the operating characteristics of normalizing flows on summarization tasks have yet to be investigated.
    
Secondly, as reported by previous studies \citep{bowman-etal-2016-generating, kingma2016improved, chen2017variational}, variational models tend to experience posterior collapse during training, which occurs when the KL term vanishes to zero, indicating that the model fails to learn meaningful latent codes. This problem becomes more severe when modeling discrete data with a strong auto-regressive decoder \citep{he2019lagging}, which is the case for Transformer-based summarization models. To resolve this issue, several solutions have been proposed, such as employing a less auto-regressive decoder network \citep{yang2017improved, semeniuta-etal-2017-hybrid, shen2018deconvolutional}, modifying the training objective \citep{zhao2017infovae, tolstikhin2018wasserstein, prokhorov-etal-2019-importance}, and proposing new training strategies \citep{kim2018semi, he2019lagging}. However, most existing work focuses on the VAE framework with Gaussian latent distribution, yet limited work considers the VED framework with normalizing flows. In particular, two questions remain unclear: (1) when the latent distribution is modeled by normalizing flows, does the posterior collapse problem still exist? (2) when posterior collapse exists, what are the appropriate strategies to achieve good summarization quality within the VED framework?

This paper introduces FlowSUM\footnote{Code is available at \url{https://github.com/yuyangstat/flowsum}.}, a normalizing flows-based VED framework for Transformer-based summarization, along with a controlled alternate aggressive training (CAAT) strategy and a refined gate mechanism to resolve the two challenging issues. Our contributions include:
\begin{enumerate} 
\item We employ normalizing flows to enrich the latent posterior distribution and integrate the latent code into Transformer-based models in a plug-and-play manner, demonstrating its effectiveness through extensive experiments. 

\item We propose a controlled alternate aggressive training strategy and a refined gate mechanism to mitigate the posterior collapse problem and improve training efficacy. 

\item Our findings suggest that FlowSUM facilitates knowledge distillation while having a negligible effect on inference time, implying normalizing flows' potential for transferring knowledge from advanced large language models.

\item We investigate the posterior collapse problem for different normalizing flows and examine how the quality of a summary is impacted by the training strategy, gate initialization, and the type and depth of normalizing flows.
\end{enumerate}

This article consists of five sections. Section \ref{sec:background} provides an overview of normalizing flows, VED, and a summary of related studies. Section \ref{sec:model} describes the proposed model architecture and the training strategies employed. Section \ref{sec:experiments} presents the experimental setup and results, and Section \ref{sec:conclusion} concludes the paper with some discussions.
    
\section{Backgrounds}\label{sec:background}

\subsection{Normalizing Flows}\label{subsec: nf}
Normalizing flows (NF) \citep{rezende2015variational} is a type of generative model that has gained popularity in recent years. The fundamental idea involves mapping a simple probability density (e.g., Gaussian) to a more complex one through a series of invertible transformations. One of the key advantages of NF is that it allows for exact likelihood evaluations, which is crucial for many applications such as density estimation \citep{papamakarios2017masked}, data generation \citep{tran2019discrete}, and variational inference \citep{kingma2016improved}. A flow-based model consists of two components: a base distribution $p_{\mathrm{u}}(\mathbf{u})$ and a transformation $f(\cdot): \mathbb{R}^D \rightarrow \mathbb{R}^D$, where $f$ must be invertible and both $f$ and $f^{-1}$ must be differentiable. Let $\mathbf{x} = f(\mathbf{u})$ where $\mathbf{u} \sim p_{\mathrm{u}}(\mathbf{u})$, then the density of $\mathbf{x}$ can be obtained via a change of variables \citep{bogachev2007measure}: 
 \begin{equation}\label{eq: nf-px}
    \begin{aligned}
    p_{\mathrm{x}}(\mathbf{x}) & = p_{\mathrm{u}}(\mathbf{u})\left|\operatorname{det} J_f(\mathbf{u})\right|^{-1} \\
    & = p_{\mathrm{u}}(f^{-1}(\mathbf{x}))\left|\operatorname{det} J_{f^{-1}}(\mathbf{x})\right|.
    \end{aligned}
    \end{equation}
In this paper, we examine several NFs, including planar flows \citep{rezende2015variational}, radial flows \citep{rezende2015variational}, Sylvester flows \citep{berg2018Sylvester}, real-valued non-volume preserving (RealNVP) transformation \citep{dinh2017density}, inverse autoregressive flow (IAF)  \citep{kingma2016improved}, rational-quadratic neural spline flows (RQNSF) \citep{durkan2019neural}, and rational-linear neural spline flows (RLNSF) \citep{dolatabadi2020invertible}. We delegate the detailed discussion of transformation and invertibility to Appendix \ref{appendix: normalizing-flows}. Throughout the paper, for each type, we compose $K$ layers of transformation $f_K \circ \cdots \circ f_1 (\cdot)$, which remains invertible and differentiable.

\subsection{Variational Encoder-Decoders}\label{subsec: ved}
Variational encoder-decoders (VEDs) \citep{zhang-etal-2016-variational-neural, serban2017hierarchical, zhou-neubig-2017-morphological, shen2018improving}, which can be seen as an extension of variational autoencoders (VAEs) \citep{kingma2014auto, rezende2014stochastic}, have been widely used to understand the conditional data generation process. Given an input $x$, the framework posits the existence of a latent variable $z \sim p(z \mid x; \phi)$, and the generation of $y$ relies on $p(y | x, z; \theta)$. With this premise, the conditional data generation can be formulated as in Eq. \ref{eq: conditional-dg}. 
\begin{equation}\label{eq: conditional-dg}
    p(y \mid x; \phi, \theta) = \int p(z \mid x; \phi) p(y \mid x, z; \theta) dz
\end{equation}
Since the marginal $p(y \mid x; \phi, \theta)$ is intractable, we employ variational inference to estimate the parameters. This involves maximizing the evidence lower bound (ELBO), a surrogate of the log-likelihood, as defined in Eq. \ref{eq: elbo-ved}. The underlying idea is to propose a parameterized distribution $q(z \mid x, y; \psi)$, known as the variational posterior, to approximate the true posterior distribution $p(z \mid x, y; \phi, \theta)$. The greater the flexibility in $q(z \mid x, y; \psi)$, the better the approximation, and the more effective the surrogate ELBO becomes. See more details in Appendix \ref{appendix: elbo-derivation}.
\begin{equation}\label{eq: elbo-ved}
\begin{aligned}
& \text{ELBO}_{\text{VED}}  \\
= & \resizebox{0.95\columnwidth}{!}{$\displaystyle\underset{q(z \mid x, y; \psi)}{\mathbb{E}}[\log p(y \mid x, z; \theta)]-\text{KL}(q(z \mid x, y; \psi) \| p(z \mid x; \phi))$}
\end{aligned}
\end{equation}
For summarization, we parameterize $p(y \mid x, z; \theta)$ as an encoder-decoder model that generates summaries conditioned on the input text and latent code. 
\subsection{Related Work}\label{sec:related-work}
\subsubsection{Transformer-based Summarization Models}

Transformer-based models equipped with pre-training and fine-tuning techniques have enjoyed significant success in many NLP tasks, including text summarization. \citet{liu-lapata-2019-text} proposed BertSUM for extractive and abstractive tasks, utilizing the pre-trained BERT encoder \citep{devlin-etal-2019-bert}. To better align the pre-trained encoder for document understanding with the decoder trained from scratch for text generation, \citet{rothe-etal-2020-leveraging} demonstrated the effectiveness of leveraging pre-trained BERT \citep{devlin-etal-2019-bert}, GPT-2 \citep{radford2019language}, and RoBERTa \citep{liu2019roberta} checkpoints to build sequence-to-sequence (S2S) models for tasks including summarization. Another approach is to address both document understanding and generation in a unified framework by first pre-training some general-purpose S2S models and then fine-tuning on downstream tasks, for instance, BART \citep{lewis-etal-2020-bart}, MASS \citep{song2019mass}, UniLM \citep{dong2019unified}, ProphetNet \citep{qi-etal-2020-prophetnet}, and T5 \citep{raffel2020exploring}. In addition, \citet{zhang2020pegasus} proposed PEGASUS with a pre-training objective tailored for abstractive summarization, achieving significant improvements across multiple datasets.

\subsubsection{Variational Summarization}

Variational summarization models come in two different flavors: unsupervised and supervised. In the unsupervised domain, researchers commonly utilize variational autoencoders in conjunction with specific control mechanisms for summary generation, as exemplified by prior work such as \citet{schumann2018unsupervised, chu2019meansum, brazinskas2020unsupervised}. In the supervised realm, there are generally two primary approaches. The first approach models the conditional probability of the target sentences $p(y \mid x)$ as in Eq. \ref{eq: conditional-dg}, whereas the second approach models the joint probability of the source and target sentences $p(x, y)$ with $\int p(z) p(x \mid z) p(y \mid z, x) dz$. Our model belongs to the first category, akin to prior studies like \citet{setiawan-etal-2020-variational, fu2020document}. In contrast, other works, including \citet{zheng2020topic, nguyen-etal-2021-enriching, zou2021topic}, adopt the second type by jointly modeling topics and sequence-to-sequence generation. Most of them assume a simple Gaussian latent prior, except for \citet{nguyen-etal-2021-enriching}, which employs normalizing flows to model neural topic models and enrich global semantics. However, they did not specify the choice of normalizing flows and how they addressed posterior collapse. To the best of our knowledge, there remains limited research on the application of normalizing flows in variational summarization models and their operating characteristics.

\section{Normalizing Flows Enhanced Summarization Model}\label{sec:model}

\subsection{FlowSUM Model Architecture} 
\begin{figure}
    \centering
    \includegraphics[width=\columnwidth]{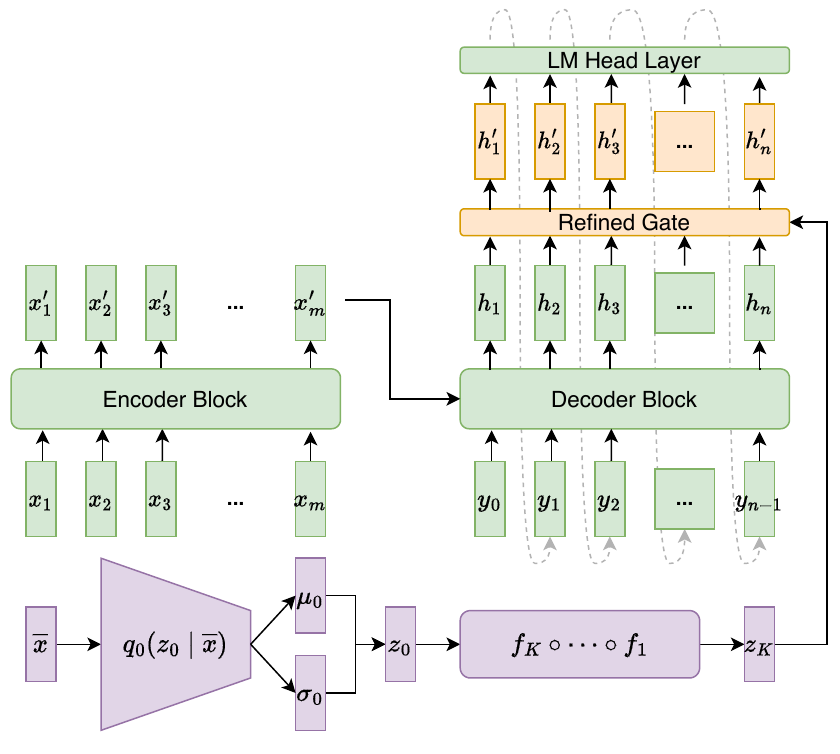}
    \caption{FlowSUM Model Architecture, including an NF latent module (in purple), a Transformer-based encoder-decoder (in green), and a refined gate mechanism (in orange)}
    \label{fig:model}
\end{figure}

As illustrated in Fig. \ref{fig:model}, FlowSUM consists of three components: an NF latent module, a Transformer-based encoder-decoder, and a refined gate mechanism. The NF latent module focuses on modeling the variational posterior $q(z \mid x, y; \psi)$, whereas the encoder-decoder, combined with the refined gate, models the conditional generation $p(y | x, z; \theta)$ with latent code. As a simplification, we assume the conditional prior $p(z \mid x; \phi)$ is a standard Gaussian as in \citet{setiawan-etal-2020-variational}. Throughout this section, let $e$ be the embedding size, $m, n$ be the length of the input source and target summary respectively, $\ell$ be the latent dimension of the NF latent module, $d$ be the dimension of the decoder's hidden states, $\{x_i\}_{i=1}^{m}$ be the input source text, $\{y_j\}_{j=1}^{n}$ be the target summary text, and $\overline{x} \in \mathbb{R}^e$ be the average embedding of the untruncated input source text\footnote{Let $V$ be the vocabulary size, $\{E_v\}_{v=1}^{V}$ be the input embeddings, and $\{b_v\}_{v=1}^V$ be the Bag-of-Words (BoW) of the input source text, then $\overline{x} = (\sum_{v=1}^V b_v E_v) / (\sum_{v=1}^V b_v ) \in \mathbb{R}^e$. In addition, when we don't truncate the input text, $b^T \mathbbm{1} = m$ holds. However, if we truncate the input due to encoder constraints, then $b^T \mathbbm{1} > m$, and the BoW vector will contain information that would otherwise have been lost.}.

\noindent\textbf{NF Latent Module.} To model the variational posterior $q(z \mid x, y; \psi)$, we follow \citet{zhou-neubig-2017-morphological} and assume all the information in $y$ is contained in $x$\footnote{See detailed discussion in Appendix \ref{appendix: discussion-q-choice}.}. Therefore, we have $q(z \mid x, y; \psi) = q(z \mid x; \psi)$, which allows us to parameterize $q(z \mid x; \psi)$ with neural networks (NNs) and normalizing flows using the amortization and reparameterization tricks \citep{kingma2014auto}. The NF latent module comprises of an inference network $q_0(\cdot)$ and a normalizing flows model. The inference network takes $\overline{x}$ as input and produces two output vectors, $\mu_0 \in \mathbb{R}^{\ell}$ and $\log (\sigma_0) \in \mathbb{R}^{\ell}$. Using the reparameterization trick, a random sample $z_0 \in \mathbb{R}^{\ell}$ is drawn from $N(\mu_0, \text{diag}(\sigma_0^2))$. Afterward, the normalizing flows model applies a sequence of $K$ invertible transformations to $z_0$ to obtain the latent code $z = z_K = f_K \circ \cdots \circ f_1(z_0) \in \mathbb{R}^{\ell}$.\footnote{The log-determinant of the Jacobian at each layer is recorded along the forward call for loss computation.} Note that when $K = 0$, the model reverts to the traditional VED framework, and we refer to this degenerated version as VEDSUM.

\noindent\textbf{Gated Transformer-based Encoder-Decoder.} Our model adopts the Transformer-based encoder-decoder. The encoder processes the input text and learns a sequence of hidden representations, and the decoder generates a summary based on the encoder's hidden states and the previously generated tokens. We incorporate the latent information into the decoder with a gate mechanism, which mixes the latent vector $z_K$ with the decoder's last layer of hidden states $\{h_j\}_{j=1}^n$. As pointed out in \citet{gu2020improving}, the saturation property of traditional gating mechanisms hinders gradient-based optimization. Therefore, following their proposal, we use a refined gate mechanism designed to allow for better gradient flow. Let $\sigma(\cdot)$ be the sigmoid function. We generate the gated fused hidden states $\{h_j^\prime\}_{j=1}^n$ as in Eq. \ref{eq: refined-gate}.
\begin{equation}\label{eq: refined-gate}
\begin{aligned}
& z_K^{\prime}=W^{z} z_K \in \mathbb{R}^d, \text{ where } W^z \in \mathbb{R}^{d \times \ell} \\
& f_j=\delta\left(W^f\left[h_j ; z_K^{\prime}\right]\right) \in \mathbb{R}^d, \text{ where } W^f \in \mathbb{R}^{d \times 2 d} \\
& r_j=\delta\left(W^r\left[h_j ; z_K^{\prime}\right]\right) \in \mathbb{R}^d, \text{ where } W^r \in \mathbb{R}^{d \times 2 d} \\
& g_j=\left(1-r_j\right) \cdot f_j^2+r_j\left(1-\left(1-f_j\right)^2\right) \in \mathbb{R}^d \\
& h_j^{\prime}=\left(1-g_j\right) \cdot h_j+g_j \cdot z_K^{\prime} \in \mathbb{R}^d
\end{aligned}
\end{equation}
Afterward, the fused hidden states are passed to a language model (LM) Head layer, where they are transformed into vectors modeling the probabilities of each word in the vocabulary.

\subsection{Training Objective} 
Traditional VEDs usually assume $q(z \mid x; \psi)$ to be a Gaussian, allowing analytical computation of the KL term in ELBO. However, in our normalizing flows-based VED, the variational posterior $q(z \mid x) = q_K (z_K \mid x)$ can be complex and hence the KL term in Eq. \ref{eq: elbo-ved} lacks an analytical form. Therefore, we rewrite the ELBO via a change of variables to enable analytical evaluation\footnote{See derivation in Appendix \ref{appendix: elbo-derivation} Eq. \ref{eq: elbo-nf-ved-derivation}.}:
\begin{equation}\label{eq: elbo-nf}
\begin{aligned}
& \text{ELBO}_{\text{NF-VED}} \\
= & \mathbb{E}_{q_0\left(z_0\right)}\left[\log p\left(y \mid x, z_K\right) + \log p\left(z_K \mid x\right)\right] \\
- & \resizebox{0.9\columnwidth}{!}{$\mathbb{E}_{q_0\left(z_0\right)}\left[\log q_0\left(z_0\right)-\sum_{k=1}^K \log \left|\operatorname{det} J_{f_k}\left(z_{k-1}\right)\right| \right]$},
\end{aligned}    
\end{equation}
where $q_0$ is $z_0$'s probability density function, a Gaussian distribution modeled by NNs, and $\operatorname{det} J_{f_k}(\cdot)$ is the determinant of $f_k$'s Jacobian.

Let $\mathcal{L}_{\text{CE}}$ denote the cross-entropy loss and $\mathcal{L}_{\text{VI}}$ denote the loss introduced by the variational latent module. Applying the idea of Monte Carlo to Eq. \ref{eq: elbo-nf}, we obtain the training objective as below. Note that $\mathcal{L}_{\text{VI}}$ is a Monte Carlo estimate of the KL divergence between the variational posterior $q_K$ and the conditional prior distribution $p(z_K \mid x)$. 
    \begin{equation}\label{eq: loss}
    \begin{aligned}
    \mathcal{L} & = \mathcal{L}_{\text {CE}} + \mathcal{L}_{\text {VI}} \\
    & = -{\textstyle\sum}_{j=1}^n \log p\left(y_j \mid\left\{x_i\right\}_{i=1}^m, z_K, y_{<j}\right) \\
    & + \log q_0\left(z_0\right)-{\textstyle\sum}_{k=1}^K \log \left|\operatorname{det} J_{f_k}\left(z_{k-1}\right)\right| \\
    & - \log p\left(z_K \mid x\right) 
    \end{aligned} 
    \end{equation}

\subsection{Mitigating Posterior Collapse}

To remedy posterior collapse, we consider two strategies, aiming to preserve the expressiveness of the latent variable and improve the overall summary quality. The first approach, called $\beta_C$-VAE \citep{prokhorov-etal-2019-importance}, replaces the KL term with $\beta|KL-C|$, where $\beta$ is a scaling factor, and $C \geq 0$ is a threshold that regulates the magnitude of the KL term. When $C > 0$, the KL term is expected to be discouraged from getting close to $0$. 

We propose the second approach, Controlled Alternate Aggressive Training (CAAT), inspired by the lagging inference strategy \citep{he2019lagging}. This strategy uses the observation that the inference network cannot accurately approximate the true posterior in the initial stages of training. As outlined in Alg. \ref{alg: cat} in Appendix \ref{appendix: caat}, CAAT comprises two stages. In the first stage, we alternately update the variational parameters and the entire parameters\footnote{In our preliminary experiments, we find that if we alternate between variational and encoder-decoder parameters, the training becomes unstable and generates NaN values. Therefore, we alternate between variational and all parameters.} for a specified number of steps. In the second stage, we train all parameters jointly, as in basic VAE training, for the remainder of the training.

\subsection{NF-enhanced Knowledge Distillation}
Normalizing flows can learn complex and multi-modal distributions \citep{papamakarios2017masked}, which makes them a promising approach for knowledge distillation tasks that involve integrating information from multiple sources \citep{hinton2015distilling}. To investigate the impact of normalizing flows on knowledge distillation, we adopt two knowledge distillation methods by \citet{shleifer2020pre}: Shrink and Fine-Tune (SFT) and Pseudo-labels (PL). SFT shrinks the teacher model and re-finetunes the shrunk model. In contrast, the PL method initializes the student model with the compressed version produced by SFT and then fine-tunes using the pseudo-labeled data generated by the teacher model. In this study, we fine-tune the model on the augmented data with both original and pseudo-labeled data, enabling it to more effectively switch between generated summaries and ground truth, thereby mitigating exposure bias.

\section{Experiments}\label{sec:experiments}

\subsection{Datasets}
We evaluate the effectiveness of FlowSUM on six public benchmark datasets\footnote{We access them through \href{https://huggingface.co/docs/datasets/index}{Hugging Face Datasets}, which provides reproducible code for processing texts and generating train/validation/test splits.}, including CNN/Daily Mail (CNN/DM) \citep{hermann2015teaching}, XSum \citep{narayan-etal-2018-dont}, Multi-News \citep{fabbri-etal-2019-multi}, arXiv, PubMed \citep{cohan-etal-2018-discourse}, and SAMSum \citep{gliwa-etal-2019-samsum}. These datasets exhibit various summary styles and lengths, and their corresponding statistics are shown in Table \ref{table: data-statistics}. Refer to Appendix \ref{appendix: datasets} for more details.

\begin{table}[h]
    \centering
    \resizebox{\columnwidth}{!}{%
    \begin{tabular}{lccc}
        \toprule
        Datasets & \makecell{Split \\(train/val/test)} & \makecell{Avg. doc\\ length} & \makecell{Avg. summary \\length}\\
        \midrule
        CNN/DM & 287113/13368/11490 & 781 & 56 \\
        Multi-News & 44972/5622/5622 & 2103 & 264\\
        arXiv & 203037/6436/6440 & 4938 & 220 \\
        PubMed & 119924/6633/6658 & 3016 & 203\\
        XSum & 204045/11332/11334 & 431 & 23 \\
        SAMSum & 14732/818/819 & 94 & 20 \\
        \bottomrule
    \end{tabular}%
    }
    \caption{\label{table: data-statistics}Statistics of Summarization Datasets.}
\end{table}
    
\subsection{Implementation Details}
We configure the inference net $q_0(z_0|\overline{x})$ to be a feedforward neural network and set the latent dimension $\ell$ to 300 and the number of NF layers $K \in \{2, 4, 6, 8\}$. For models that use $\beta_C$-VAE, we set $\beta=1$ and $C = 0.1$, and for those using CAAT, we conduct one epoch of aggressive training with $n_{alt} = 15$ and two epochs of non-aggressive training. See more details in Appendix \ref{appendix: implementation-details}. 

\subsection{Baselines}
We use BART \citep{lewis-etal-2020-bart} and BERT2BERT \citep{rothe-etal-2020-leveraging} as two backbone models. We refer to the PL knowledge distilled FlowSUM as FlowSUM-PLKD. Our comparison involves the following baselines: PG+Cov \citep{see-etal-2017-get}, BERT2BERT \citep{rothe-etal-2020-leveraging}, BERTSUM \citep{liu-lapata-2019-text}, BART \citep{lewis-etal-2020-bart}, PEGASUS \citep{zhang2020pegasus}, VHTM \citep{fu2020document}, TAS \citep{zheng2020topic}, and PEGASUS+Flow-NTM \citep{nguyen-etal-2021-enriching}. See Appendix \ref{appendix: baseline-models} for more detailed descriptions.

\subsection{Results}

\subsubsection{Automatic Evaluation}
We evaluate the generated summary quality using ROUGE scores \citep{lin-2004-rouge} and BERTScore \citep{zhang2020bertscore}\footnote{We obtain both metrics using \href{https://huggingface.co/docs/evaluate/index}{Hugging Face Evaluate} and report the $F_1$ scores.}. Specifically, we utilize the overlap of unigrams and bigrams (ROUGE-1 and ROUGE-2) to evaluate the informativeness, and the longest common subsequence (ROUGE-L) for fluency.  Moreover, we report BERTScore, which gauges semantic similarity based on contextual embeddings. Furthermore, we present rep-w \citep{fu2021theoretical}\footnote{rep-w is calculated as the proportion of the current token that appears in the previous $w$ tokens. Refer to Appendix \ref{appendix: repetition-measures} for the detailed definition.} and the average length of summaries to gain a better understanding of the quality.

We compare the proposed model against baseline models in ROUGE scores in Tables \ref{table: comp_with_baselines_cnndm} and \ref{table: comp_with_baselines_xsum}. On CNN/DM, FlowSUM (BERT2BERT) greatly outperforms BERT2BERT, whereas VEDSUM adds noise to the model and leads to a decrease in performance. With the BART backbone, FlowSUM achieves an absolute improvement over the BART model with +0.48, +0.08, and +0.75 in R-1, 2, and L scores, respectively.
However, on XSum, the variational models do not perform well when the gold summaries involve only one sentence. VEDSUM leads to a significant decrease in performance, whereas with FlowSUM, the decrease in ROUGE scores is less severe, leading to +0.12, -0.15, and -0.25 in R-1, 2, and L scores, respectively.

Table \ref{table: comp_with_baselines_all_datasets} uses BART as the backbone and compares BART, VEDSUM, and FlowSUM across all datasets. Overall, variational models produce summaries of superior quality for datasets with long summaries, such as CNN/DM, Multi-News, arXiv, and PubMed, and FlowSUM further enhances the performance beyond VEDSUM. However, when it comes to datasets featuring short summaries such as XSum and SAMSum, the variational component markedly diminishes the model performance. We hypothesize that brief summaries may be more susceptible to disturbances and are more prone to being affected by noise. Nevertheless, incorporating NF modules alleviates these reductions and accomplishes comparable outcomes. Furthermore, we observe that both variational models tend to generate lengthier summaries, while FlowSUM exhibits fewer issues with repetition compared to VEDSUM.

\begin{table}
   \centering
   \resizebox{\columnwidth}{!}{%
   \begin{tabular}{lccc}
        \toprule
        \multirow{2}{*}{Model} & \multicolumn{3}{c}{ROUGE $\uparrow$} \\
        \cline{2-4} 
        & 1 & 2 & L \\
        \midrule
        PG+Cov \citep{see-etal-2017-get} & 39.53 & 17.28 & 36.38 \\ 
        BERT2BERT \citep{rothe-etal-2020-leveraging} & 41.28 & 18.69 & 38.09 \\         
        BERTSUM \citep{liu-lapata-2019-text} & 42.13 & 19.60 & 39.18 \\
        BART \citep{lewis-etal-2020-bart} & 44.16 & 21.28 & 40.90  \\ 
        PEGASUS \citep{zhang2020pegasus} & 44.17 & 21.47 & 41.11 \\ 
        \midrule
        VHTM \citep{fu2020document} & 40.57 & 18.05 & 37.18 \\
        TAS \citep{zheng2020topic} & 44.38 & 21.19 & 41.33 \\
        PEGASUS+NTM \citep{nguyen-etal-2021-enriching} & 44.52 & \textbf{21.95} & 41.39 \\
        \midrule
        \midrule
        VEDSUM (BERT2BERT) & 40.89 &	18.28 & 37.95 \\
        FlowSUM (BERT2BERT) & 41.51 & 18.81 & 38.56 \\
        \midrule
        VEDSUM (BART) & 44.36 & 21.09 & 41.37 \\
        FlowSUM (BART)  & \textbf{44.64} & 21.36 & \textbf{41.65} \\
        FlowSUM-PLKD (BART) & 44.59 & 21.49 & 41.59 \\
        \bottomrule
    \end{tabular}%
    }
    \caption{\label{table: comp_with_baselines_cnndm}Comparison with baselines on CNN/DM.}
\end{table}

\begin{table}
   \centering
   \resizebox{\columnwidth}{!}{%
   \begin{tabular}{lccc}
        \toprule
        \multirow{2}{*}{Model} & \multicolumn{3}{c}{ROUGE $\uparrow$} \\
        \cline{2-4} 
        & 1 & 2 & L \\
        \midrule
        PG+Cov \citep{see-etal-2017-get} & 28.10 & 8.02 & 21.72 \\ 
        BERTSUM \citep{liu-lapata-2019-text} & 38.81 & 16.50 & 31.27 \\
        BART \citep{lewis-etal-2020-bart} & 45.14 & 22.27 & 37.25 \\ 
        PEGASUS \citep{zhang2020pegasus} & 47.21 & 24.56 & 39.25\\ 
        \midrule
        TAS \citep{zheng2020topic} & 44.63 & 21.62 & 36.77 \\
        PEGASUS+NTM \citep{nguyen-etal-2021-enriching} & \textbf{49.57} & \textbf{25.08} & \textbf{41.81} \\
        \midrule
        \midrule
        VEDSUM (BART) & 43.62 & 20.27 & 35.06\\
        FlowSUM (BART)  & 45.26 & 22.12 & 37.00\\
        FlowSUM-PLKD (BART) & 45.54 & 22.67 & 37.38\\
        \bottomrule
    \end{tabular}%
    }
    \caption{\label{table: comp_with_baselines_xsum}Comparison with baselines on XSum.}
\end{table}

\begin{table}[t]
   \centering
   \resizebox{\linewidth}{!}{%
   \begin{tabular}{lcccc}
        \toprule
        Model & \makecell{ROUGE $\uparrow$\\1/2/L} & \makecell{BERT-\\Score} $\uparrow$ & \makecell{rep-w} $\downarrow$ & Length\\
        \midrule
        \multicolumn{5}{c}{\textbf{CNN/DM}} \\
        \midrule
        BART & 44.16/21.28/40.90 & 89.40 & \textbf{8.31} & 84.11 \\
        VEDSUM & 44.34/21.09/41.37 & 89.20 & 8.43 & 88.63 \\ 
        FlowSUM & \textbf{44.64}/\textbf{21.36}/\textbf{41.65} & \textbf{89.46} & 8.43 & 92.24 \\ 

        \midrule
        \multicolumn{5}{c}{\textbf{Multi-News}} \\
        \midrule
        BART & 42.56/15.34/36.67 & 86.69 & \textbf{9.76} & 133.42 \\ 
        VEDSUM & 43.91/16.68/38.10 & 87.04 & 9.95 & 128.79 \\ 
        FlowSUM & \textbf{44.42}/\textbf{17.01}/\textbf{38.36} & \textbf{87.09} & 9.91 & 128.87 \\ 

        \midrule
        \multicolumn{5}{c}{\textbf{arXiv}} \\
        \midrule
        BART & 42.55/15.92/37.89 & 85.35 & 17.23 & 130.68 \\ 
        VEDSUM & 43.05/\textbf{16.34}/38.26 & 85.44 & 16.63 & 130.92 \\ 
        FlowSUM & \textbf{43.11}/16.26/\textbf{38.31} & \textbf{85.45} & \textbf{16.55} & 132.88 \\ 

        \midrule
        \multicolumn{5}{c}{\textbf{PubMed}} \\
        \midrule
        BART & 41.57/16.72/36.94 & 84.65 & 13.26 & 136.10 \\ 
        VEDSUM & 44.21/19.20/39.32 & 85.07 & 12.76 & 138.70 \\ 
        FlowSUM & \textbf{44.55}/\textbf{19.50}/\textbf{39.59} & \textbf{85.16} & \textbf{12.59} & 138.09 \\ 

        \midrule
        \multicolumn{5}{c}{\textbf{XSum}} \\
        \midrule
        BART & 45.14/\textbf{22.27}/\textbf{37.25} & \textbf{92.16} & \textbf{4.63} & 25.54 \\ 
        VEDSUM & 43.62/20.27/35.06 & 91.75 & 5.96 & 31.22 \\ 
        FlowSUM & \textbf{45.26}/22.12/37.00 & 92.13 & 4.95 & 28.71 \\ 
        \midrule

        \multicolumn{5}{c}{\textbf{SAMSum}} \\
        \midrule
        BART & \textbf{53.16}/28.19/\textbf{49.03} & \textbf{92.68} & 6.71 & 30.00 \\ 
        VEDSUM & 51.91/26.74/47.41 & 92.40 & 7.53 & 30.92 \\ 
        FlowSUM & 53.13/\textbf{28.49}/49.00 & 92.67 & \textbf{6.59} & 29.77 \\ 
        \bottomrule
    \end{tabular}%
    }
  \caption{\label{table: comp_with_baselines_all_datasets}Comparison of BART, VEDSUM (BART), and FlowSUM (BART) on all six benchmarks.}
\end{table}

\subsubsection{On NF-enhanced Knowledge Distillation}
We use PEGASUS as the teacher model to generate pseudo-labels on the CNN/DM training set. In this study, we explore the effects of knowledge distillation on BART and DistilBART, a shrunken version of BART. We examine two variations of DistilBART: dBART-6-6, which replicates 6 layers\footnote{The 0, 2, 4, 7, 9, and 11th layer.} of the BART encoder and decoder, and dBART-12-3, which duplicates all layers of the BART encoder and 3 layers\footnote{The 0, 6, and 11th layer.} of the decoder.

Table \ref{table: kd_on_bart_cnndm} presents the impact of the PL approach on the original BART model. Training the BART model on augmented data worsens the performance compared to training on the original data. In contrast, VEDSUM-PLKD achieves improvements in all three ROUGE scores, and FlowSUM-PLKD with RQNSF achieves the highest R-2 score, albeit with some sacrifice in R-1 and R-L\footnote{This can be explained by the teacher model's worse performance in these two metrics.}. However, planar flows appear to be unsuitable for knowledge distillation via PL. To better understand FlowSUM-PLKD, we visualize the latent distribution (see Appendix \ref{appendix: latent-visualization}) and demonstrate how the NF's ability to capture multi-modality could account for its impressive performance.

\begin{table}
   \centering
   \resizebox{\columnwidth}{!}{%
   \begin{tabular}{lccccccc}
        \toprule
        \multirow{2}{*}{Model} & \multicolumn{3}{c}{ROUGE $\uparrow$} & \multirow{2}{*}{\makecell{BERT-\\Score} $\uparrow$} & \multirow{2}{*}{Length}\\
        \cline{2-4}
        & 1 & 2 & L \\
        \midrule
        BART & 44.16 & 21.28 & 40.90 & 89.40 & 84.11 \\ 
        VEDSUM & 44.34 & 21.09 & 41.37 & 89.20 & 88.63 \\ 
        FlowSUM (Planar) & 44.62 & 21.32 & 41.64 & 89.20 & 90.78 \\ 
        FlowSUM (RQNSF) & \textbf{44.64} & 21.36 & \textbf{41.65} & 89.46 & 92.24 \\ 
        
        \midrule
        PEGASUS & 44.17 & 21.47 & 41.11 & \textbf{89.52} & 77.84 \\ 
        \midrule
        
        BART-PLKD & 42.83 & 20.16 & 39.98 & 89.04 & 100.52 \\ 
        VEDSUM-PLKD & 44.45 & 21.25 & 41.45 & 89.41 & 93.42 \\ 
        FlowSUM-PLKD (Planar) & 44.19 & 21.03 & 41.15 & 89.34 & 92.38 \\ 
        FlowSUM-PLKD (RQNSF) & 44.59 & \textbf{21.48} & 41.59 & 89.47 & 84.75 \\ 
        \bottomrule
    \end{tabular}%
    }
    \caption{\label{table: kd_on_bart_cnndm}PL Knowledge Distillation on BART on CNN/DM.}
\end{table}

Table \ref{table: kd_on_distilbart_cnndm} investigates the two DistilBART variants with RQNSF. With FlowSUM, both variants achieve improvements, suggesting that NF is beneficial for the SFT approach. Previous experiments from \citet{shleifer2020pre} showed that PL performed worse than SFT on CNN/DM. However, our experiments reveal that the NF latent module unleashes the potential of PL. When trained on augmented data, FlowSUM-PLKD (dBART-6-6) achieves R-1/2/L improvements of 0.92/0.47/1.01 over dBART-6-6, and FlowSUM-PLKD (dBART-12-3) achieves improvements of 0.66/0.49/0.63 over dBART-12-3, much more than the SFT approach. Furthermore, FlowSUM does not introduce additional computational burden at inference, and the time cost is primarily related to the length of the generated summaries.


\begin{table}
   \centering
   \resizebox{\columnwidth}{!}{%
   \begin{tabular}{lccccc}
        \toprule
        Model & \makecell{ROUGE $\uparrow$ \\1/2/L} & \makecell{BERT-\\Score} $\uparrow$ & Length & \makecell{\# Params \\(MM)} & \makecell{Inference \\Time (MS)} $\downarrow$\\
        \midrule
        \multicolumn{6}{c}{\textbf{dBART-6-6}} \\
        \midrule
        dBART-6-6 & 42.78/20.24/39.72 & 88.98 & 67.42 & 230 & 170.5 \\ 
        FlowSUM & 43.41/20.33/40.41 & 89.18 & 91.25 & 238 & 234.9\\ 
        FlowSUM-PLKD & \textbf{43.70}/\textbf{20.71}/\textbf{40.73} & \textbf{89.24} & 91.10 & 238 & 239.7 \\ 
        \midrule
        \multicolumn{6}{c}{\textbf{dBART-12-3}} \\
        \midrule
        dBART-12-3 & 43.39/20.57/40.44 & 89.20 & 85.48 & 255 & 199.6 \\ 
        FlowSUM & 43.53/20.61/40.59 & 89.28 & 83.74 & 263 & 190.7\\ 
        FlowSUM-PLKD & \textbf{44.05}/\textbf{21.06}/\textbf{41.07} & \textbf{89.37} & 84.48 & 263 & 200.4 \\ 
        \bottomrule
    \end{tabular}%
    }
    \caption{\label{table: kd_on_distilbart_cnndm}Knowledge Distillation on DistilBART on CNN/DM.}
\end{table}

\subsubsection{Analysis on NF Types and Depth}
We investigate the effect of NF types and the number of NF layers on the Multi-News dataset\footnote{We choose Multi-News due to its smaller size, enabling us to conduct experiments with reduced computational cost.}. Table \ref{table: effect_nf_types_multinews} explores the effect of NF types. Simple flows like Planar and Radial yield inferior performance compared to the VAE counterpart, whereas more complex flows tend to achieve greater improvements. Overall, IAF and RQNSF emerge as the best-performing NF types.

Table \ref{table: effect_nf_num_layers_multinews} delves further into IAF and RQNSF, investigating the effect of NF depth. The findings indicate that adding more layers does not always lead to improved performance. We hypothesize that when the encoder-decoder model is well-trained, the increased complexity of the NF module may introduce more noise, outweighing the benefits of better latent modeling and subsequently worsening the summary quality.

\begin{table}
   \centering
   \resizebox{\columnwidth}{!}{%
   \begin{tabular}{lcccc}
        \toprule
        Model & \makecell{ROUGE $\uparrow$ \\1/2/L} & \makecell{BERT-\\Score} $\uparrow$ & \makecell{rep-w} $\downarrow$ & Length\\
        \midrule
        BART & 42.56/15.35/36.67 & 86.69 & \textbf{9.76} & 133.42 \\ 
        VEDSUM & 43.91/16.68/38.10 & 87.04 & 9.95 & 128.79 \\ 
        FlowSUM (Planar) & 43.85/16.61/37.97 & 87.03 & 10.04 & 128.84 \\ 
        FlowSUM (Radial) & 43.84/16.68/37.98 & 87.04 & 9.92 & 128.72 \\ 
        FlowSUM (Sylvester) & 44.18/16.71/38.15 & 87.08 & 9.80 & 128.76 \\ 
        FlowSUM (RealNVP) & 44.19/16.64/38.15 & 87.05 & 9.81 & 128.76 \\ 
        FlowSUM (IAF) &\textbf{44.42}/\textbf{17.01}/\textbf{38.36} & \textbf{87.09} & 9.91 & 128.87 \\ 
        FlowSUM (RLNSF) & 44.25/16.86/38.14 & 87.06 & 9.80 & 128.80 \\ 
        FlowSUM (RQNSF) & 44.31/16.98/38.27 & 87.07 & 9.91 & 128.81 \\ 
        \bottomrule
    \end{tabular}%
    }
    \caption{\label{table: effect_nf_types_multinews}Effect of NF Types on Multi-News.}
\end{table}

\begin{table}
   \centering
   \resizebox{\columnwidth}{!}{%
   \begin{tabular}{lcccc}
        \toprule
        Model & \makecell{ROUGE $\uparrow$ \\1/2/L} & \makecell{BERT-\\Score} $\uparrow$ & \makecell{rep-w} $\downarrow$ & Length\\
        \midrule
        FlowSUM (IAF-4) & 44.30/\textbf{17.03}/38.22 & 87.05 & \textbf{9.82} & 128.81 \\ 
        FlowSUM (IAF-6) & \textbf{44.42}/17.01/\textbf{38.36} & \textbf{87.09} & 9.91 & 128.87 \\ 
        FlowSUM (IAF-8) & 44.18/16.90/38.16 & 87.04 & 9.88 & 128.84  \\
        \midrule
        FlowSUM (RQNSF-2) & 44.15/16.88/38.20 & 87.04 & 9.94 & 128.83 \\ 
        FlowSUM (RQNSF-4) & \textbf{44.31}/\textbf{16.98}/\textbf{38.27} & \textbf{87.07} & 9.91 & 128.81 \\ 
        FlowSUM (RQNSF-6) & 44.15/16.88/38.18 & 87.06 & \textbf{9.87} & 128.92 \\ 
        \bottomrule
    \end{tabular}%
    }
    \caption{\label{table: effect_nf_num_layers_multinews}Effect of Number of NF Layers on Multi-News.}
\end{table}

\subsubsection{Analysis on Training Strategies}
We implement standard VAE training, $\beta_C$-VAE, and CAAT on VEDSUM and FlowSUM models, and we evaluate their effectiveness with different types of normalizing flows. Table \ref{table: effect-of-training-strategies} shows that VEDSUM and FlowSUM models with residual flows, including planar, radial, and Sylvester flows, suffer from posterior collapse, whereas those with more complex flows do not. Moreover, applying $\beta_C$-VAE to VEDSUM and FlowSUM models with residual flows does not effectively mitigate posterior collapse but even exacerbates the issue. Furthermore, for models with planar, RealNVP, and IAF flows, training with $\beta_C$-VAE worsens ROUGE scores, while for radial and Sylvester flows, it improves performance. Notably, the two neural spline flows are not impacted by $\beta_C$-VAE training. 

Concerning CAAT, we note that applying it to treat severe posterior collapses such as VEDSUM and FlowSUM with residual flows can cause instability in training while producing NaN values. Hence, it is only effective for models with KL divergence that is not close to zero. Nonetheless, when applicable, CAAT enhances the quality of summaries, particularly when utilized with the top-performing NFs, namely IAF and RQNSF.

\begin{table}
\centering
  \begin{threeparttable}[b]
  \resizebox{\columnwidth}{!}{%
   \begin{tabular}{llcccccccc}
        \toprule
        \multirow{2}{*}{Model} &  \multirow{2}{*}{Training} & \multicolumn{3}{c}{ROUGE $\uparrow$} & \multirow{2}{*}{\makecell{KL \\Divergence}} \\
        \cline{3-5}
        & & 1 & 2 & L \\
        \midrule
        VEDSUM & standard & 43.91 & 16.68 & 38.10 & 0.0117 \\ 
        VEDSUM & $\beta_C$-VAE & 43.78 & 16.54 & 37.96 & 0.0082\\ 
        \midrule
        FlowSUM (Planar) & standard & 43.85 & 16.61 & 37.97 & 0.2719 \\ 
        FlowSUM (Planar) &$\beta_C$-VAE & 43.68 & 16.47 & 37.85 & 0.1815\\ 
        \midrule
        FlowSUM (Radial) & standard & 43.63 & 16.37 & 37.82 & 0.0121\\ 
        FlowSUM (Radial) & $\beta_C$-VAE & 43.84 & 16.68 & 37.98 & 0.0096\\ 
        \midrule
        FlowSUM (Sylvester) & standard & 43.68 & 16.51 & 37.87 & 0.0841\\ 
        FlowSUM (Sylvester) & $\beta_C$-VAE & 44.18 & 16.71 & 38.15 & 0.0348 \\
        \midrule
        FlowSUM (RealNVP) & standard & 44.19 & 16.64 & 38.15 & 4.7986\\ 
        FlowSUM (RealNVP) & $\beta_C$-VAE & 43.71 & 16.54 & 37.85 & 7.8938 \\ 
        FlowSUM (RealNVP) & CAAT & 44.12 & 16.82 & 38.11 & 5.2107 \\ 
        \midrule
        FlowSUM (IAF) & standard & 43.87 & 16.62 & 37.97 & 3.9146 \\ 
        FlowSUM (IAF) & $\beta_C$-VAE & 43.81 & 16.58 & 37.91 & 3.9128 \\
        FlowSUM (IAF) & CAAT & 44.30 & 17.03 & 38.22 & 2.1108 \\ 
        \midrule
        FlowSUM (RLNSF) & standard & 44.25 & 16.86 & 38.14 & 104.9667 \\ 
        FlowSUM (RLNSF) & $\beta_C$-VAE & 44.25 & 16.86 & 38.14 & 104.9667 \\
        FlowSUM (RLNSF) & CAAT & 44.14 & 16.82 & 38.05 & 95.3774 \\ 
        \midrule
        FlowSUM (RQNSF) & standard & 44.18 & 16.76 & 38.18 & 127.8106\\ 
        FlowSUM (RQNSF) & $\beta_C$-VAE & 44.18 & 16.76 & 38.18 & 127.8106 \\
        FlowSUM (RQNSF) & CAAT & 44.31 & 16.98 & 38.27 & 107.0794 \\ 
        \bottomrule
    \end{tabular}%
    }
     \begin{tablenotes}
       \item[a] \scriptsize{VEDSUM and FlowSUM with radial flows have no CAAT results as \\the training is unstable and generates NaN values.}
     \end{tablenotes}
  \end{threeparttable}
  \caption{\label{table: effect-of-training-strategies}Effect of Training Strategies.}
\end{table}

In addition, we explore the impact of gate score initialization. The standard method initializes gating weights with small deviations from zero, resulting in an initial gate score close to 0.5. In contrast, the near-zero initialization method initializes gating weights such that the resulting gate score is approximately 0.05. Our experiments using FlowSUM (BERT2BERT) with RQNSF as the base model reveal that CAAT + Standard Gate Score Initialization yields the best results and the most stable training process, as illustrated in Table \ref{table: effect-of-caat-and-gate} and Figures \ref{fig: train-comp} to \ref{fig: train-details} in Appendix \ref{appendix: exp-training-gateinit}. This suggests that by setting a large initial gate score and forcing the model to learn from the NF latent module, we can better capture latent code information.

\begin{table}
\centering
  \resizebox{\columnwidth}{!}{%
   \begin{tabular}{lcccccccc}
        \toprule
        \multirow{2}{*}{Training} & \multirow{2}{*}{Gate Initialization} & \multicolumn{3}{c}{ROUGE $\uparrow$} \\
        \cline{3-5}
        & & 1 & 2 & L \\
        \midrule
        standard & standard & 40.82 & 18.29 & 37.92 \\ 
        standard & near-zero & 40.98 & 18.36 & 38.09 \\ 
        CAAT & standard & \textbf{41.51} & \textbf{18.81} & \textbf{38.56} \\ 
        CAAT & near-zero & 41.13 & 18.57 & 38.21 \\ 
        \bottomrule
    \end{tabular}%
    }
    \caption{\label{table: effect-of-caat-and-gate}Effect of CAAT and Gate Initialization.}
\end{table}

\section{Conclusions and Discussions}\label{sec:conclusion}
This paper introduces FlowSUM, a normalizing flows-based Variational Encoder-Decoder (VED) framework for text summarization. It outperforms a leading non-latent model across multiple datasets. This enhanced performance is attributed to the flexible posterior distributions provided by normalizing flows. We also analyze the operating characteristics and the posterior collapse problem of normalizing flows and propose an effective training strategy for complex flows. Moreover, we demonstrate that incorporating normalizing flows is highly effective for knowledge distillation with minimal impact on inference time.

FlowSUM illustrates the advantages of incorporating flexible latent modeling. Considering the remarkable achievements of Latent Diffusion Models (LDMs) in generating images \citep{rombach2022high}, adopting LDMs for capturing latent representation may produce comparable or even superior outcomes in text summarization. In this scenario, the gating mechanism may not be an appropriate choice.  A direct correlation between the latent vector and the target text may be more suitable for executing the diffusion process. Enhancing the architecture to leverage diffusion models could be a potential avenue for future research.

\section*{Limitations}
FlowSUM has demonstrated excellent results on datasets with long summaries. However, its performance on short-summary datasets like XSum and SAMSum has been unsatisfactory. The underlying cause could be attributed to suboptimal hyperparameter tuning or the incompatibility of FlowSUM with short summaries. Additional investigations are needed to identify the root cause.

Furthermore, we did not fine-tune the hyper-parameters of the normalizing flows model, such as the latent dimension, the number of bins in spline coupling layers, and the neural network in IAF, RealNVP, RLNSF, and RQNSF. Moreover, we opted for a small batch size due to memory limitations. Adjusting these hyperparameters could potentially enhance the model’s performance.

Due to limited computational resources, we utilized BART and BERT2BERT as the backbone models instead of newer architectures. Further research may focus on verifying the effectiveness of FlowSUM on more advanced structures.

\section*{Ethics Statement}
Our research entailed developing a new text summarization framework. Although no private data were utilized, we acknowledge the potential societal impacts of our work. Therefore, we adhered to pertinent ethical guidelines and implemented rigorous procedures to guarantee the accuracy of our results.

\section*{Acknowledgements}
This work was supported in part by NSF grant DMS-1952539 and NIH grants R01AG069895, R01AG065636, R01AG074858, U01AG073079.

\bibliography{anthology,custom}
\bibliographystyle{acl_natbib}

\appendix
\section{Controlled Alternate Aggressive Training (CAAT)}\label{appendix: caat}
\begin{algorithm}[h]
    \caption{Controlled Alternate Aggressive Training (CAAT)} 
    \label{alg: cat}
    \textbf{Input:} number of aggressive training steps $n_{agg}$; maximum number of training steps $n_{max}$; number of alternating steps $n_{alt}$.
    \begin{algorithmic}[1]
        \State $\boldsymbol{\theta}, \boldsymbol{\psi} \leftarrow$ Initialize encoder-decoder parameters and variational parameters respectively
        \For {$i=1,2,\cdots, n_{agg}$}
        \State $\mathbf{X} \leftarrow$ Random data minibatch
        \If {$i \mod n_{alt} = 0$}
        \State Compute  $\boldsymbol{g}_{\boldsymbol{\theta}, \boldsymbol{\psi}} \leftarrow \nabla_{\boldsymbol{\psi}, \boldsymbol{\theta}} \mathcal{L}(\mathbf{X} ; \boldsymbol{\theta}, \boldsymbol{\psi})$
        \State Update $\boldsymbol{\theta}, \boldsymbol{\psi}$ using gradients $\boldsymbol{g}_{\boldsymbol{\theta}, \boldsymbol{\psi}}$
        \Else
        \State Compute $\boldsymbol{g}_\psi \leftarrow \nabla_{\boldsymbol{\psi}} \mathcal{L}(\mathbf{X} ; \boldsymbol{\theta},\boldsymbol{\psi})$
        \State Update $\boldsymbol{\psi}$ using graidents $\boldsymbol{g}_{\boldsymbol{\psi}}$
        \EndIf
        \EndFor

        \For {$i=n_{agg},n_{agg} + 1,\cdots, n_{max}$}
        \State $\mathbf{X} \leftarrow$ Random data minibatch
        \State Compute $\boldsymbol{g}_{\boldsymbol{\theta}, \boldsymbol{\psi}} \leftarrow \nabla_{\boldsymbol{\psi}, \boldsymbol{\theta}} \mathcal{L}(\mathbf{X} ; \boldsymbol{\theta}, \boldsymbol{\psi})$
        \State Update $\boldsymbol{\theta}, \boldsymbol{\psi}$ using gradients $\boldsymbol{g}_{\boldsymbol{\theta}, \boldsymbol{\psi}}$
        \If {early stopping criterion is met}
        \State \textbf{break}
        \EndIf
        \EndFor
    \end{algorithmic} 
\end{algorithm}

Another advantage of the controlled alternate aggressive training (CAAT) strategy is that it provides us with more control. It is commonly assumed that allowing the model more freedom to learn, even if the NF latent module is not helpful, will not harm performance. However, our experiments suggest that this assumption does not hold, particularly for short-summary datasets where the model will not learn on its own to avoid hurting the original performance. The CAAT strategy allows us to effectively freeze the encoder-decoder parameters by setting $n_{agg}$ and $n_{alt}$ to large values, ensuring that when the nf module is unhelpful, it will not significantly harm performance.

\section{Deeper Dive into the Evidence Lower Bound (ELBO)}\label{appendix: elbo-derivation}
Within the VED framework, the conditional data generation process can be expressed as follows:
$$p(y \mid x; \phi, \theta) = \int p(z \mid x; \phi) p(y \mid x, z; \theta) dz.$$
The subsequent challenge revolves around parameter estimation. Typically, the conditional latent prior is assumed as $p(z \mid x; \phi) = N(0, I)$ for simplification (hence eliminating the $\phi$ parameter). Despite this, the likelihood $p(y \mid x; \theta)$ remains computationally intractable to evaluate. Variational inference tackles this issue by introducing a variational distribution $q(z \mid x, y; \psi)$ from a specific parametric family, aiming to approximate the actual posterior $p(z \mid x, y)$. Here, $\theta$ denotes the model parameters, and $\psi$ refers to the variational parameters. Instead of attempting to estimate $\theta$ solely through maximizing the challenging log-likelihood, the approach involves joint estimation of both $\theta$ and $\psi$ by optimizing the ELBO.

Examining Eq. \ref{eq: elbo-ved-derivation} and \ref{eq: elbo-ved-derivation_2}, it's evident that the ELBO represents a lower bound of the log-likelihood. Moreover, a smaller value of $\mathrm{KL}(q(z \mid x, y) \| p(z \mid x, y))$ indicates a closer alignment between the variational posterior and the true posterior, thereby bringing the ELBO closer to the log-likelihood. This insight propels the adoption of normalizing flows to model a flexible family of variational posterior.

\begin{footnotesize}
\begin{equation}\label{eq: elbo-ved-derivation}
\begin{aligned}
& \mathrm{KL}(q(z \mid x, y) \| p(z \mid x, y)) \\
= & \mathbb{E}_{q(z \mid x, y)}[\log q(z \mid x, y)]-\mathbb{E}_{q(z \mid x, y)}\left[\log \frac{p(z, x, y)}{p(x, y)}\right] \\
= & \mathbb{E}_{q(z \mid x, y)}[\log q(z \mid x, y)] \\ 
& -\mathbb{E}_{q(z \mid x, y)}\left[\log \frac{p(z, x, y)}{p(x, z)} \cdot \frac{p(x, z)}{p(x)} \cdot \frac{p(x)}{p(x, y)}\right] \\
= & \mathbb{E}_{q(z \mid x, y)}[\log q(z \mid x, y)]-\mathbb{E}_{q(z \mid x, y)}[\log p(y \mid x, z)] \\
& - \mathbb{E}_{q(z \mid x, y)}[\log p(z \mid x)] + \mathbb{E}_{q(z \mid x, y)}[\log p(y \mid x)] \\
= & K L(q(z \mid x, y) \| p(z \mid x)) - \mathbb{E}_{q(z \mid x, y)}[\log p(y \mid x, z)] \\ 
& + \mathbb{E}_{q(z \mid x, y)}[\log p(y \mid x)] \\
\geqslant & 0 \\ \\
\end{aligned}
\end{equation}
\end{footnotesize}

\begin{footnotesize}
\begin{equation}\label{eq: elbo-ved-derivation_2}
\begin{aligned}
& \text{ELBO}_{\text{VED}} \\
= & \mathbb{E}_{q(z \mid x, y)}[\log p(y \mid x, z)] - K L(q(z \mid x, y) \| p(z \mid x)) \\
= & \log p(y \mid x) - \mathrm{KL}(q(z \mid x, y) \| p(z \mid x, y)) \\
\leq & \log p(y \mid x) 
\end{aligned}
\end{equation}
\end{footnotesize}

\begin{footnotesize}
\begin{equation}\label{eq: elbo-nf-ved-derivation}
\begin{aligned}
& \text{ELBO}_{\text{NF-VED}} \\
= & \mathbb{E}_{q(z \mid x)}[\log p(y \mid x, z)]+\mathbb{E}_{q(z \mid x)} \log p(z \mid x) \\
& -\mathbb{E}_{q(z \mid x)}[\log q(z \mid x)]\\
= & \mathbb{E}_{q_0\left(z_0\right)}\left[\log p\left(y \mid x, z_K\right)+\log p\left(z_K \mid x\right)\right] \\
& - \mathbb{E}_{q_0\left(z_0\right)}\left[\log q_K\left(z_K\right) \right] \\
= & \mathbb{E}_{q_0\left(z_0\right)}\left[\log p\left(y \mid x, z_K\right) + \log p\left(z_K \mid x\right)\right] \\
& - \mathbb{E}_{q_0\left(z_0\right)}\left[\log q_0\left(z_0\right)-\sum_{k=1}^K \log \left|\operatorname{det} J_{f_k}\left(z_{k-1}\right)\right| \right],
\end{aligned}    
\end{equation}
\end{footnotesize}
\noindent where $q_0$ and $q_K$ are the probability density function for $z_0$ and $z_K$ respectively.

\section{Discussion on $q(z \mid x, y) = q(z \mid x)$}\label{appendix: discussion-q-choice}
we choose to assume $q(z \mid x, y) = q(z \mid x)$ for the following reasons. Firstly, this assumption is grounded in the nature of summarization, where $y$ can be viewed as a condensed form of $x$ and hence it is sensible to assume all the information in $y$ is contained in $x$. Secondly, as evidenced by \citet{zhang-etal-2016-variational-neural}, it is plausible to condition the posterior on both $x$ and $y$. However, their approach suffers from difficulties during prediction. In prediction, the target text $y$ is not accessible, making it hard to sample from $q(z \mid x, y)$. \citet{zhang-etal-2016-variational-neural} suggests taking the prior’s mean as the latent code, but in our paper, the prior is a Gaussian whereas the posterior is a complex distribution modeled by normalizing flows, and taking such a strategy would diminish the benefit of using normalizing flows. Thirdly, it has been shown empirically by \citet{eikema2019auto} that by restricting the conditioning of the posterior to $x$ alone, their model achieves higher accuracy. Therefore, we consider $q(z \mid x, y) = q(z \mid x)$ as our modeling strategy.

\section{Repetition Measures}\label{appendix: repetition-measures}
    Let $s$ represent the sentences in a result set $\mathcal{D}$, $|s|$ be the number of tokens in $s$, $s_t$ be the $t$th token, and $s_{i:j}$ be the sub-sequence of $s$ from the $i$th token to the $j$th token. The rep-w \citep{fu2021theoretical} is then defined by Equation \ref{eq: repw}.
    \begin{equation}\label{eq: repw}
        \textrm{rep-w} = \frac{1}{|\mathcal{D}|} \sum_{s \in \mathcal{D}} \frac{1}{|s|} \sum_{t=2}^{|s|} \mathbbm{1}\left[s_{t} \in s_{\max\left(t-w, 1\right): t-1}\right]
    \end{equation}

\section{Datasets}\label{appendix: datasets}
\noindent\textbf{CNN/Daily Mail} \citep{hermann2015teaching} consists of 312,085 online news articles, with one article paired with a multi-sentence summary. We use the non-anonymized version as in \citet{see-etal-2017-get} and follow the text processing\footnote{We update the data loading script following \url{https://github.com/facebookresearch/fairseq/issues/1401}.} in \citet{lewis-etal-2020-bart}.

\noindent\textbf{XSum} \citep{narayan-etal-2018-dont} contains 227k BBC articles, each summarized in a single sentence.

\noindent\textbf{Multi-News} \citep{fabbri-etal-2019-multi} is a multi-document dataset comprising 56k pairs of news articles and multi-sentence summaries.

\noindent\textbf{arXiv, PubMed} \citep{cohan-etal-2018-discourse} are two scientific paper document datasets from arXiv.org (113k) and PubMed (215k). Each pair consists of a scientific article's body document and its abstract.

\noindent\textbf{SAMSum} \citep{gliwa-etal-2019-samsum} includes 16k conversations annotated with summaries by linguists. Unlike structured texts, the information in dialogues is scattered across different speakers' utterances, increasing the summarization difficulty.

\section{Baseline Models}\label{appendix: baseline-models}
\noindent\textbf{PG+Cov} \citep{see-etal-2017-get} is a pointer-generator (PG) network supplemented with a coverage mechanism that addresses the Out-Of-Vocabulary problem and minimizes word repetition.

\noindent\textbf{BERT2BERT} \citep{rothe-etal-2020-leveraging} initializes both the encoder and the decoder with the pre-trained BERT checkpoints and adds cross-attention layers.

\noindent\textbf{BERTSUM} \citep{liu-lapata-2019-text} builds on top of BERT and applies a fine-tuning scheduler to better align the encoder and the decoder.

\noindent\textbf{BART} \citep{lewis-etal-2020-bart} is a pretrained denoising autoencoder with the standard sequence-to-sequence Transformer architecture. In this paper, we use BART as the encoder-decoder backbone.

\noindent\textbf{PEGASUS} \citep{zhang2020pegasus} is a large Transformer-based S2S model,  pre-trained on massive text data using a self-supervised objective called gap sentence generation, designed for abstractive summarization.

\noindent\textbf{VHTM} \citep{fu2020document} is a variational hierarchical model built on the PG network. It models the topic proportion vector with isotropic Gaussian and fuses in topic information at diverse granularity levels.

\noindent\textbf{TAS} \citep{zheng2020topic} is a topic-guided Transformer-based S2S model that injects the topic-word matrix into the LMHead layer and jointly trains the NTM and encoder-decoder model.

\noindent\textbf{PEGASUS+Flow-NTM} \citep{nguyen-etal-2021-enriching} is a topic-aware model built on PEGASUS. It utilizes a Flow-based NTM and a contextualized gating mechanism to integrate topic information into the encoder and the decoder.





\section{Implementation Details}\label{appendix: implementation-details}
\subsection{NF Latent Module}\label{appendix: nf-latent-module}
We configure the inference net $q(z_0|\overline{x})$ to be a feedforward neural network with three hidden layers of dimension $\in \{300, 600\}$, Tanh activations, and a 0.1 dropout rate. We set the latent dimension $\ell$ to 300 and the number of NF layers $\in \{2, 4, 6, 8\}$. For spline coupling layers (RLNSF and RQNSF), we set the number of bins to 4, the bound to 3.0, the split dimension to $\ell/2$, and the neural network to have two hidden layers with the dimension $\ell$. For RealNVP, the split dimension is $\ell/2$, and the neural network has one hidden layer with a dimension of $10\ell$. For IAF, the neural network features one hidden layer of the dimension $3\ell + 1$. Moreover, we set $\beta=1$ and $C = 0.1$ for models that use $\beta_C$-VAE, and for models that use CAAT, we conduct one epoch of aggressive training with $n_{alt} = 15$, followed by two epochs of non-aggressive training.

\subsection{Optimization}\label{appendix: optimization}
We train the models using the Adam optimizer \citep{kingma2015adam} with $\beta_1 = 0.9, \beta_2 = 0.999$, and $\epsilon = 10^{-8}$. The initial learning rate is $5\times 10^{-5}$. We employ a linear learning rate scheduler that increases the learning rate from 0 to the initial learning rate during the warmup stage and decreases it from the initial learning rate to 0 after the warmup stage. We also apply the gradient clipping technique with a maximum gradient norm of 1.0. Furthermore, we terminate the training early when the perplexity fails to improve for eight or sixteen consecutive evaluation calls.

\subsection{Model Hyper Parameters}\label{appendix: hyper-param}
Table \ref{table: model_hyper-params} provides the hyper-parameters for the models discussed in Table \ref{table: comp_with_baselines_all_datasets} - \ref{table: effect_nf_types_multinews}, for the sake of reproducibility. To ensure fair comparisons, unless otherwise specified, the VEDSUM models typically employ the same set of hyper-parameters as their FlowSUM counterparts, except with standard training and no NF layers applied. Additionally, the models in Table \ref{table: effect_nf_num_layers_multinews} have the same hyper-parameters as those in Table \ref{table: effect_nf_types_multinews}, except for the number of NF layers used. Lastly, in Table \ref{table: effect-of-training-strategies}, all FlowSUM models use 4 NF layers and the same set of hyper-parameters as those in Table \ref{table: effect_nf_types_multinews} but vary in their training strategies.

\begin{table*}[t]
  \begin{threeparttable}[b]
   \centering
   \resizebox{\textwidth}{!}{%
   \begin{tabular}{lcccccccccc}
        \toprule
        \multicolumn{11}{c}{\textbf{FlowSUM in Table \ref{table: comp_with_baselines_all_datasets}}} \\
        \midrule
        Dataset & \makecell{Number of \\epochs} & \makecell{Number of \\aggressive epochs} & \makecell{Batch \\size} & \makecell{Inference net \\hidden dim} & \makecell{NF type} & \makecell{Number of \\NF layers} & \makecell{Beam \\size} & \makecell{Length \\penalty} & \makecell{Max input \\tokens} & \makecell{Max target \\tokens} \\
        \midrule
        CNN/Daily Mail & 3 & 1 & 8 & 300 & RQNSF & 4 & 4 & 2.0 & 1024 & 128 \\
        Multi-News & 3 & 1 & 8 & 600 & IAF & 6 & 4 & 2.0 & 1024 & 128 \\
        arXiv & 4 & 1 & 16 & 600 & RQNSF & 4 & 4 & 2.0 & 1024 & 142 \\
        PubMed & 4 & 1 & 16 & 600 & RQNSF & 6 & 4 & 2.0 & 1024 & 142 \\
        XSum & 3 & 1 & 8 & 600 & RQNSF & 4 & 6 & 0.5 & 1024 & 62\\
        SAMSum & 12 & 12 & 8 & 600 & RQNSF & 4 & 6 & 1.0 & 1024 & 62 \\
        \midrule
        \midrule
        \multicolumn{11}{c}{\textbf{Models in Table \ref{table: kd_on_bart_cnndm}}} \\
        \midrule
        Model & \makecell{Number of \\epochs} & \makecell{Number of \\aggressive epochs} & \makecell{Batch \\size} & \makecell{Inference net \\hidden dim} & \makecell{NF type} & \makecell{Number of \\NF layers} & \makecell{Beam \\size} & \makecell{Length \\penalty} & \makecell{Max input \\tokens} & \makecell{Max target \\tokens} \\
        \midrule
        VEDSUM & 3 & 0 & 8 & 600 & -\tnote{a} & - & 4 & 2.0 & 1024 & 128 \\
        FlowSUM (Planar) & 3 & 0 & 8 & 600 & Planar & 4 & 4 & 2.0 & 1024 & 128 \\
        FlowSUM (RQNSF) & 3 & 1 & 8 & 300 & RQNSF & 4 & 4 & 2.0 & 1024 & 128 \\
        BART-PLKD & 3 & 0 & 8 & - & - & - & 4 & 2.0 & 1024 & 128 \\
        VEDSUM-PLKD & 3 & 0 & 8 & 600 & - & - & 4 & 2.0 & 1024 & 128 \\
        FlowSUM-PLKD (Planar) & 3 & 0 & 8 & 600 & Planar & 4 & 4 & 2.0 & 1024 & 128 \\
        FlowSUM-PLKD (RQNSF) & 3 & 1 & 8 & 300 & RQNSF & 4 & 4 & 2.0 & 1024 & 128 \\
        \midrule
        \midrule
        \multicolumn{11}{c}{\textbf{Models in Table \ref{table: kd_on_distilbart_cnndm}}} \\
        \midrule
        Model & \makecell{Number of \\epochs} & \makecell{Number of \\aggressive epochs} & \makecell{Batch \\size} & \makecell{Inference net \\hidden dim} & \makecell{NF type} & \makecell{Number of \\NF layers} & \makecell{Beam \\size} & \makecell{Length \\penalty} & \makecell{Max input \\tokens} & \makecell{Max target \\tokens} \\
        \midrule
        \multicolumn{11}{c}{dBART-6-6} \\
        \midrule
        FlowSUM & 3 & 1 & 8 & 300 & RQNSF & 4 & 4 & 2.0 & 1024 & 128 \\
        FlowSUM-PLKD & 3 & 1 & 8 & 300 & RQNSF & 4 & 4 & 2.0 & 1024 & 128 \\
        \midrule
        \multicolumn{11}{c}{dBART-12-3} \\
        \midrule
        FlowSUM & 3 & 1 & 8 & 300 & RQNSF & 4 & 4 & 2.0 & 1024 & 128 \\
        FlowSUM-PLKD & 3 & 1 & 8 & 300 & RQNSF & 4 & 4 & 2.0 & 1024 & 128 \\
        \midrule
        \midrule
        \multicolumn{11}{c}{\textbf{Models in Table \ref{table: effect_nf_types_multinews}}} \\
        \midrule
        Model & \makecell{Number of \\epochs} & \makecell{Training \\strategy} & \makecell{Batch \\size} & \makecell{Inference net \\hidden dim} & \makecell{NF type} & \makecell{Number of \\NF layers} & \makecell{Beam \\size} & \makecell{Length \\penalty} & \makecell{Max input \\tokens} & \makecell{Max target \\tokens} \\
        \midrule
        FlowSUM (Planar) & 3 & standard & 8 & 600 & Planar & 4 & 4 & 2.0 & 1024 & 128 \\
        FlowSUM (Radial) & 3 & $\beta_C$-VAE & 8 & 600 & Radial & 4 & 4 & 2.0 & 1024 & 128 \\
        FlowSUM (Sylvester) & 3 & $\beta_C$-VAE & 8 & 600 & Sylvester & 4 & 4 & 2.0 & 1024 & 128 \\
        FlowSUM (RealNVP) & 3 & standard & 8 & 600 & RealNVP & 4 & 4 & 2.0 & 1024 & 128 \\
        FlowSUM (IAF) & 3 & 1/3 CAAT\tnote{b} & 8 & 600 & IAF & 6 & 4 & 2.0 & 1024 & 128 \\
        FlowSUM (RLNSF) & 3 & $\beta_C$-VAE & 8 & 600 & RLNSF & 4 & 4 & 2.0 & 1024 & 128 \\
        FlowSUM (RQNSF) & 3 & 1/3 CAAT & 8 & 600 & RQNSF & 4 & 4 & 2.0 & 1024 & 128 \\
        \bottomrule
    \end{tabular}%
    }
     \begin{tablenotes}
       \item[a] \scriptsize{"-" means not applicable.}
       \item[b] \scriptsize{1/3 CAAT: aggressive training for 1 epoch and non-aggressive training for 2 epochs.}
     \end{tablenotes}
  \end{threeparttable}
  \caption{\label{table: model_hyper-params}Model Hyper-parameters.}
\end{table*}

\section{Experiments on Training Strategies and Gate Initialization}\label{appendix: exp-training-gateinit}
\begin{figure}
\centering
    \begin{subfigure}{0.5\columnwidth}
      \centering
      \includegraphics[width=\linewidth]{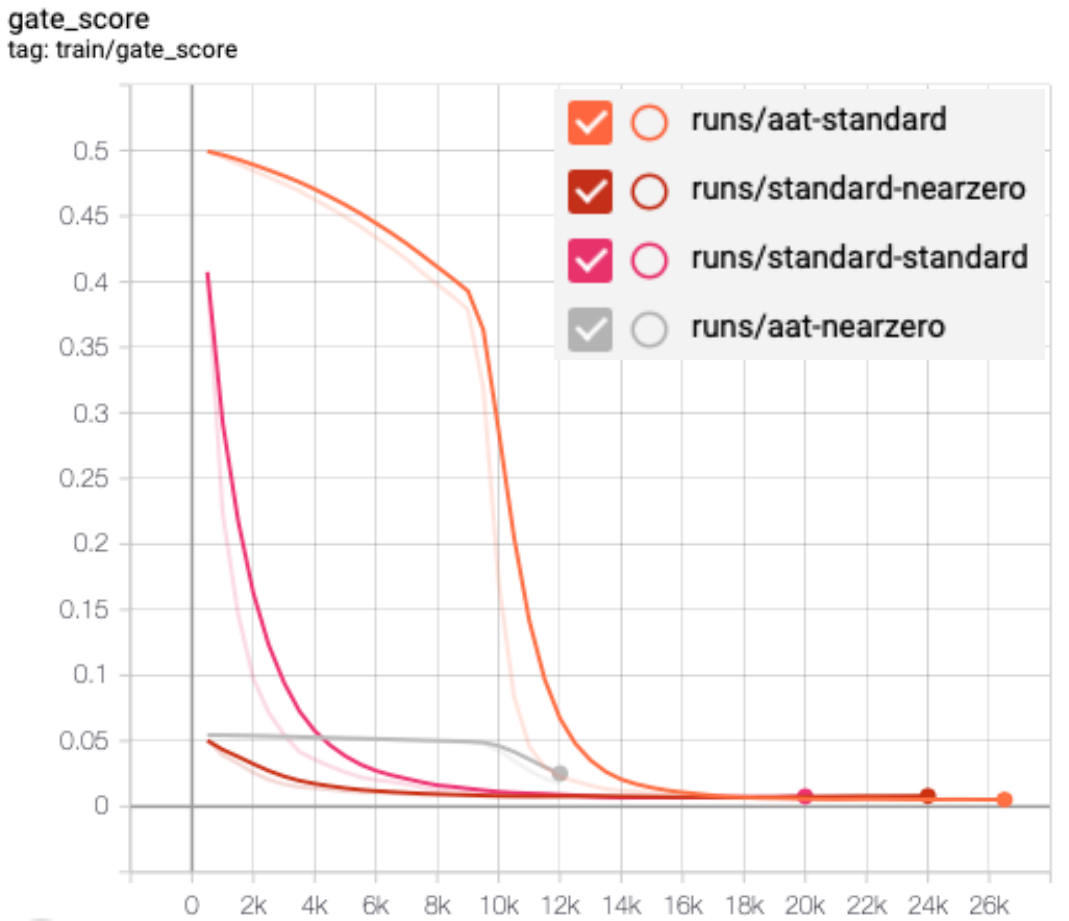}
      \caption{Gate Score}
      \label{fig: gate-score}
    \end{subfigure}%
    \begin{subfigure}{0.5\columnwidth}
      \centering
      \includegraphics[width=\linewidth]{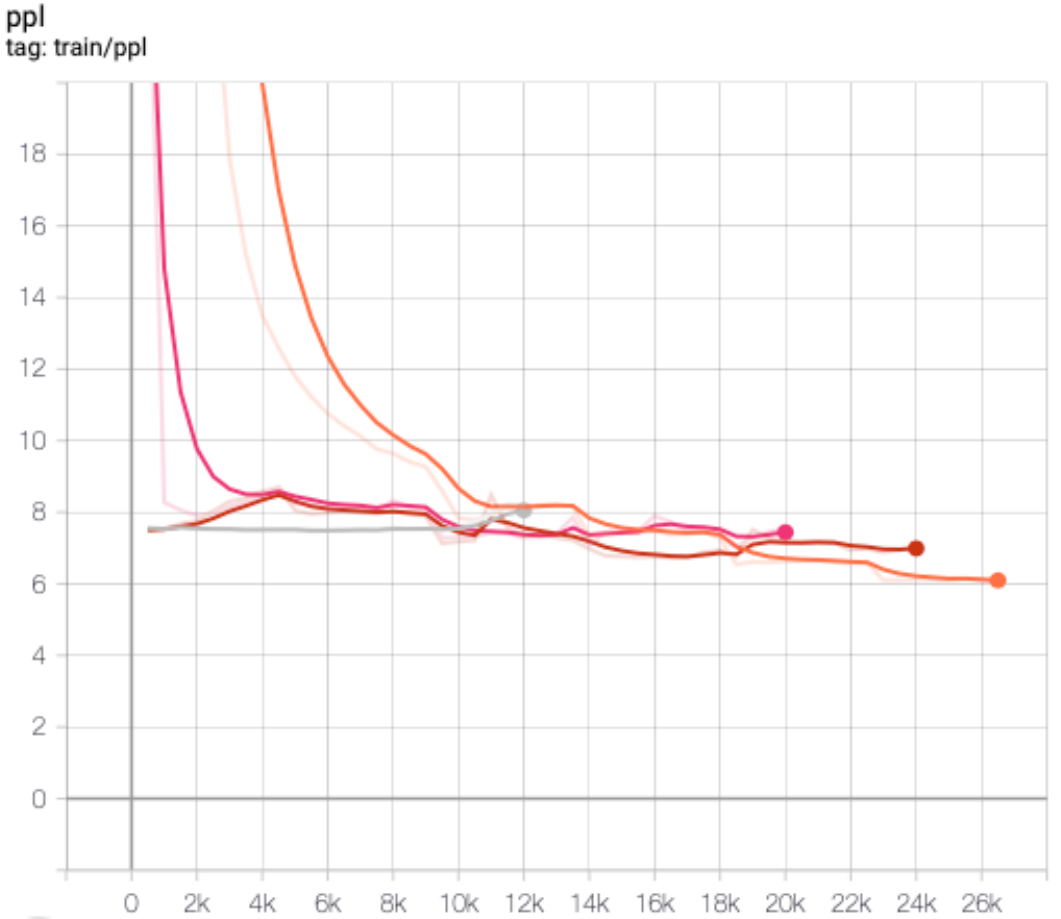}
      \caption{Training Perplexity}
      \label{fig: train-ppl}
    \end{subfigure}
    \begin{subfigure}{0.5\columnwidth}
      \centering
      \includegraphics[width=\linewidth]{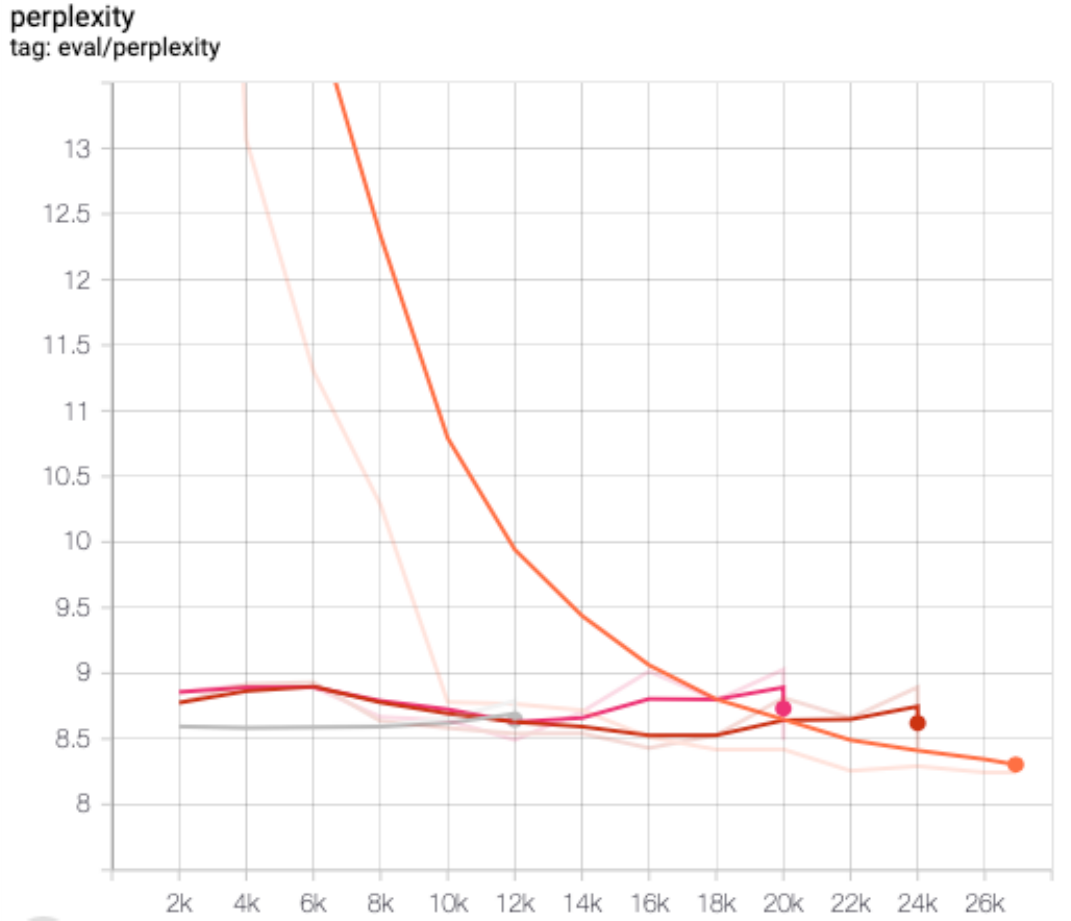}
      \caption{Evaluation Perplexity}
      \label{fig: eval-ppl}
    \end{subfigure}
\caption{Comparison of training strategies and gate initialization.}        
\label{fig: train-comp}
\end{figure}

\begin{figure}
\centering
    \begin{subfigure}{0.5\columnwidth}
      \centering
      \includegraphics[width=\linewidth]{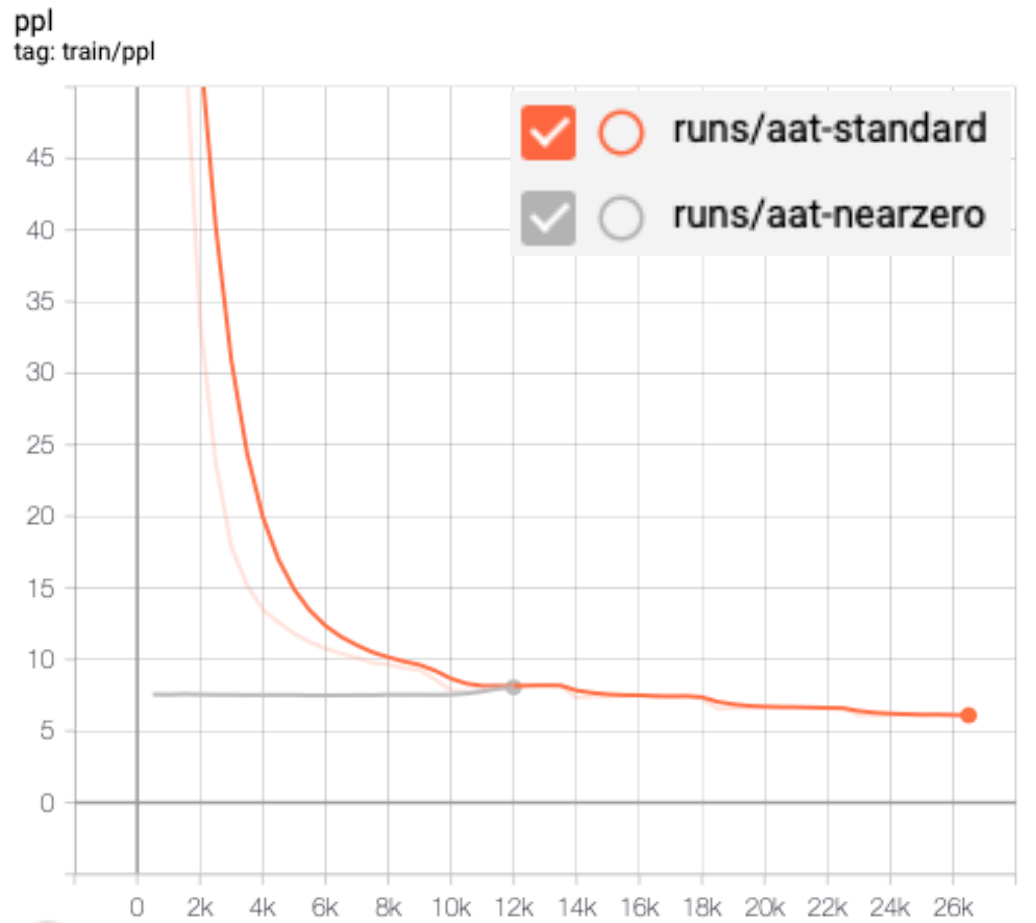}
      \caption{CAAT}
      \label{fig: train-ppl-caat}
    \end{subfigure}%
    \begin{subfigure}{0.5\columnwidth}
      \centering
      \includegraphics[width=\linewidth]{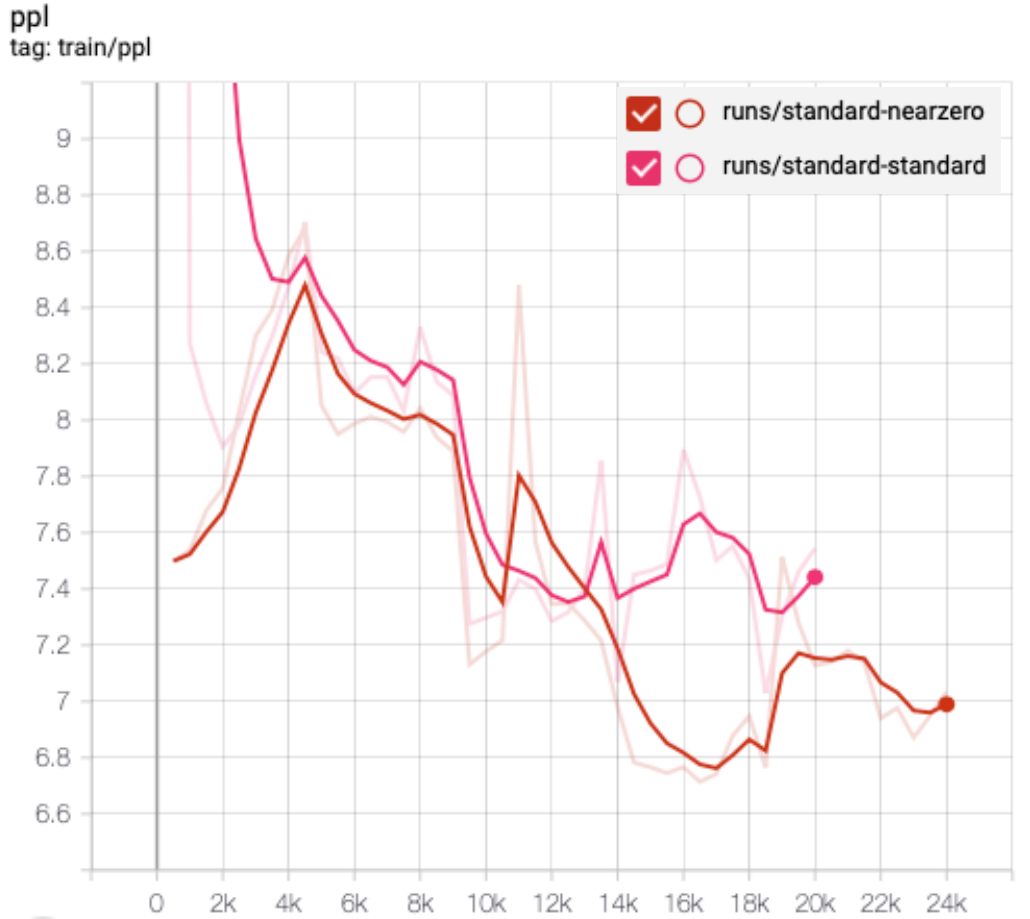}
      \caption{Standard Training}
      \label{fig: train-ppl-standard}
    \end{subfigure}
\caption{A closer look at the training process: CAAT vs. Standard Training.}        
\label{fig: train-details}
\end{figure}

The training curves for the methods in Table \ref{table: effect-of-caat-and-gate} are illustrated in Figure \ref{fig: train-comp}. The plot demonstrates that the gate score decreases gradually and remains high during aggressive training when CAAT is combined with standard initialization. This combination compels the model to utilize the latent code information effectively. Moreover, as presented in Figure \ref{fig: eval-ppl}, even though CAAT combined with standard initialization starts with a high perplexity, it achieves a lower perplexity level than other approaches by the end. By examining the training procedure in detail, Figure \ref{fig: train-details} further indicates that CAAT contributes to greater training stability than standard training.

\section{Visualization of Latent Distribution}\label{appendix: latent-visualization}
\begin{figure}
    \centering
    \includegraphics[width=\columnwidth]{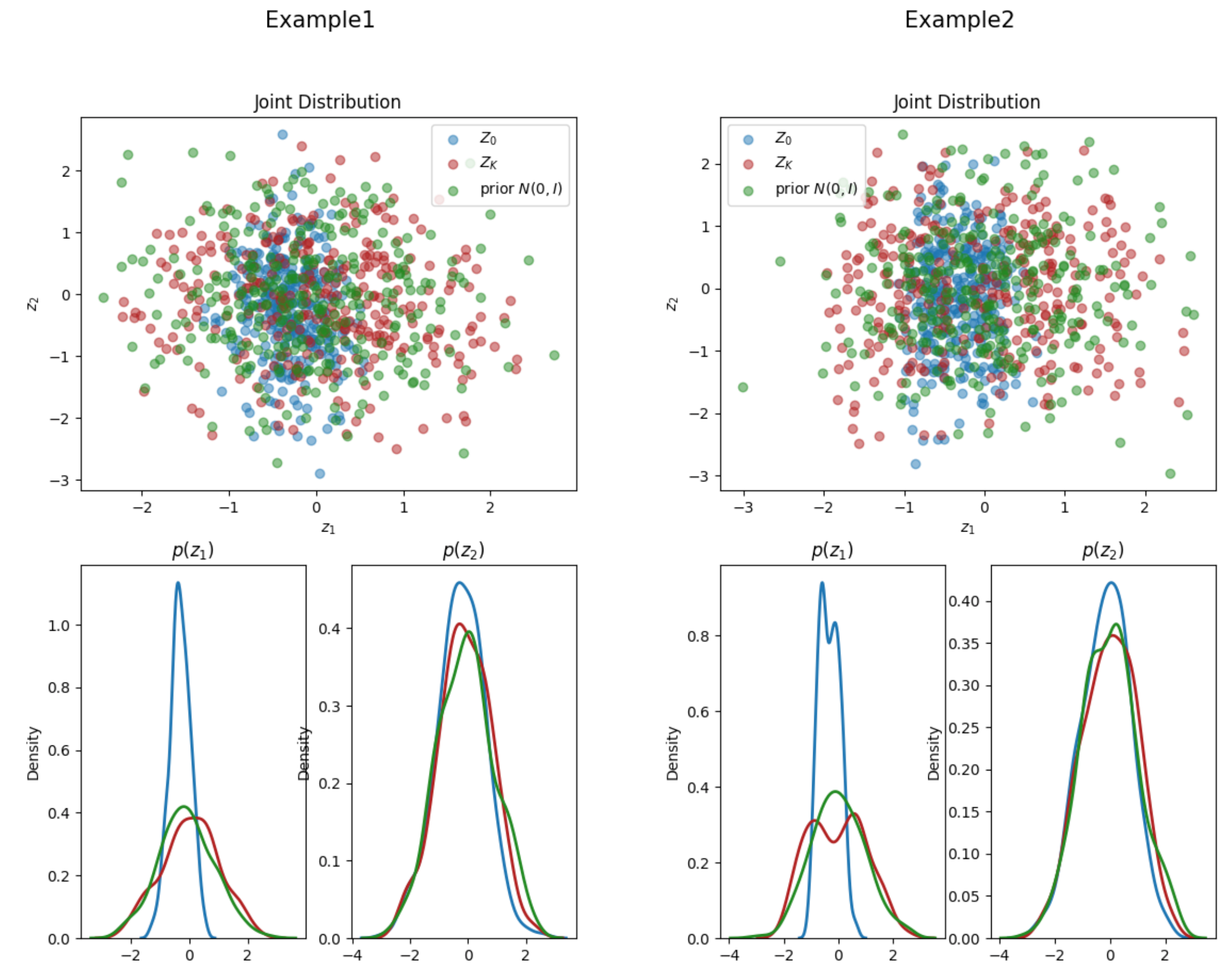}
    \caption{Visualization of the first two dimensions of $z_0$, $z_K$, and $N(0, I)$ by FlowSUM-PLKD on CNN/DM. The right sub-figure demonstrates a clear bi-modality.}
    \label{fig: latent_visual_cnndm}
\end{figure}

\begin{figure}
    \centering
    \includegraphics[width=\columnwidth]{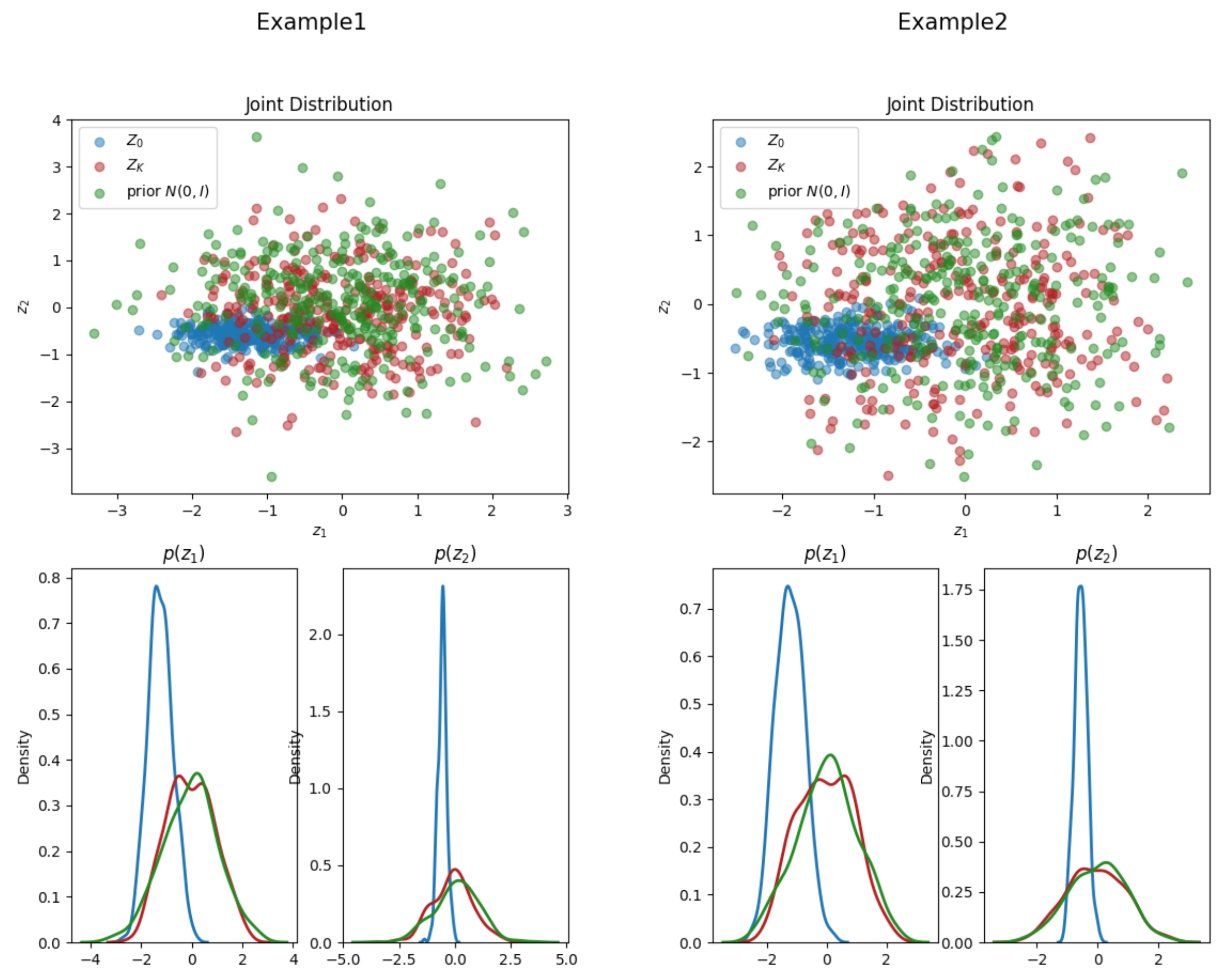}
    \caption{Visualization of the first two dimensions of $z_0$, $z_K$, and $N(0, I)$ by FlowSUM-PLKD on XSum. Both sub-figures demonstrate distinct multi-modal patterns.}
    \label{fig: latent_visual_xsum}
\end{figure}

To gain a better understanding of how normalizing flows contribute to knowledge distillation, we selected several examples from the CNN/Daily Mail and XSum datasets and visualized the resulting latent distribution generated by the FlowSUM-PLKD model, as shown in Figure \ref{fig: latent_visual_cnndm} and \ref{fig: latent_visual_xsum}. For both cases, the transformed latent code $z_K$ exhibited a highly flexible distribution. Notably, in the CNN/Daily Mail example, the first dimension of the second example demonstrated a clear bi-modal distribution, indicating the model's ability to capture information from multiple sources. Similarly, in the XSum dataset examples, we observed distinct multi-modal patterns.

\section{Normalizing Flows}\label{appendix: normalizing-flows}
\textbf{Planar flow} Proposed by \citet{rezende2015variational}, the planar flow can be expressed as in Eq. \ref{eq: planar-appendix}. It applies contractions or expansions in the direction perpendicular to the hyperplane $\mathbf{w}^{\top} \mathbf{z}+b=0$. Its Jacobian determinant can be computed in time $\mathcal{O}(D)$ as in Eq. \ref{eq: planar-jacobian-det-appendix}, using the matrix determinant lemma. In addition, we need to note that this flow is not invertible for all values of $\mathbf{u}$ and $\mathbf{w}$. When the derivative of the activation function $h^{\prime}(\cdot)$ is positive and bounded from above, $\mathbf{w}^{\top} \mathbf{u}>-\frac{1}{\sup _x h^{\prime}(x)}$ is sufficient to ensure invertibility\footnote{In our code, we perform a transformation on $\mathbf{u}: \mathbf{u} \leftarrow \mathbf{u} + \left[\log \left(1+\exp \left(\mathbf{w}^{\top} \mathbf{u}\right)\right)-1-\mathbf{w}^{\top} \mathbf{u}\right] \cdot \frac{\mathbf{w}}{\mathbf{w}^{\top} \mathbf{w}}$ and restrict the activation $h(\cdot)$ to be one of leakyrelu, relu, and tanh to meet this condition.}.

\begin{equation}\label{eq: planar-appendix}
f(\mathbf{z})=\mathbf{z}+\mathbf{u} h\left(\mathbf{w}^{\top} \mathbf{z}+b\right),
\end{equation}
\begin{equation}\label{eq: planar-jacobian-det-appendix}
\operatorname{det} J = 1 + h^{\prime}\left(\mathbf{w}^{\top} \mathbf{z}+b\right) \mathbf{w}^{\top} \mathbf{u}
\end{equation}
where $\{\mathbf{u}, \mathbf{w} \in \mathbb{R}^D, b \in \mathbb{R} \}$ are free parameters and $h(\cdot)$ is a smooth element-wise non-linear activation function with derivative $h^{\prime}(\cdot)$.

\textbf{Radial flow} The radial flow \citep{tabak2013family, rezende2015variational} takes the form of Eq. \ref{eq: radial-appendix}. It applies radial contractions and expansions around a reference point. Similar to the planar flow, we can apply the matrix determinant lemma to calculate the Jacobian determinant in $\mathcal{O}(D)$ time, as in Eq. \ref{eq: radial-jacobian-det-appendix}. To guarantee invertibility, we usually require $\beta>-\alpha$\footnote{In our code, we perform a transformation on $\beta: \beta \leftarrow-\alpha+\log \left(1+e^\beta\right)$ to guarantee invertibility.}.

\begin{equation}\label{eq: radial-appendix}
    f(\mathbf{z})=\mathbf{z}+\beta h(\alpha, r)\left(\mathbf{z}-\mathbf{z}_0\right),
\end{equation}
\begin{equation}\label{eq: radial-jacobian-det-appendix}
\operatorname{det} J = \left(1+\frac{\alpha \beta}{h^2}\right)\left(1 + \beta h\right)^{D-1}
\end{equation}
where $\mathbf{z}_0 \in \mathbb{R}^D$ is the reference point, $\beta \in \mathbb{R}, \alpha \in \mathbb{R}^{+}$ are free parameters, $r=\left\|z-z_0\right\|$ is the norm of $z-z_0$, and $h(\alpha, r)=\frac{1}{\alpha+r}$.

\textbf{Sylvester flow} The Sylvester flows \citep{berg2018Sylvester} generalize the planar flows to have $M$ hidden units, as in Eq. \ref{eq: Sylvester-appendix}. To achieve better computational efficiency, \citet{berg2018Sylvester} proposes the parameterization as in Eq. \ref{eq: Sylvester-2-appendix}, with which the Jacobian determinnant reduces to Eq. \ref{eq: Sylvester-jacobian-det-appendix} and can be computed in $\mathcal{O}(M)$. Similar to the planar flows, when $h^{\prime}(\cdot)$ is positive and bounded from above, $\tilde{\mathbf{R}}_{ii} \mathbf{R}_{ii} > -\frac{1}{\sup _x h^{\prime}(x)}$ for all $i \in\{1, \ldots, D\}$ is sufficient to ensure invertibility.

\begin{equation}\label{eq: Sylvester-appendix}
f(\mathbf{z}) = \mathbf{z}+\mathbf{U} h\left(\mathbf{W}^{\top} \mathbf{z}+\mathbf{b}\right),
\end{equation}
where $\{\mathbf{U} \in \mathbb{R}^{D \times M}, \mathbf{W} \in \mathbb{R}^{D \times M}, \mathbf{b} \in \mathbb{R}^M \}$ are the free parameters and $h(\cdot)$ is an element-wise activation function.

\begin{equation}\label{eq: Sylvester-2-appendix}
f(\mathbf{z}) = \mathbf{z}+\mathbf{Q} \mathbf{R} h\left(\tilde{\mathbf{R}} \mathbf{Q}^T \mathbf{z}+\mathbf{b}\right), 
\end{equation}
\begin{equation}\label{eq: Sylvester-jacobian-det-appendix}
\resizebox{\columnwidth}{!}{$\operatorname{det} J  = \operatorname{det} \left( \mathbf{I}_M + \operatorname{diag} \left( h^{\prime} \left( \tilde{\mathbf{R}} \mathbf{Q}^T \mathbf{z}+\mathbf{b}\right)\right) \tilde{\mathbf{R}} \mathbf{R}\right)$}
\end{equation}
where $\mathbf{R}$ and $\tilde{\mathbf{R}}$ are upper triangular $M \times M$ matrices, and $\mathbf{Q}=\left(\mathrm{q}_1 \ldots \mathrm{q}_M\right)$ consists of an orthonormal set of vectors.

\textbf{Autoregressive Flows} The masked autoregressive flow (MAF) \citep{papamakarios2017masked} was motivated by MADE \citep{germain2015made}, which is an autoregressive model for density estimation. MAF generalizes the conditional distribution to be Gaussian and generates data in a recursive way as in Eq. \ref{eq: maf-appendix}. Given a data point $\mathbf{x}$, the inverse transformation can be performed in parallel as in Eq. \ref{eq: maf-inverse-appendix}. The Jacobian of the inverse transformation is lower-triangular by design due to the autoregressive structure, hence its absolute determinant can be expressed as in Eq. \ref{eq: maf-jacobian-appendix}. The set of functions $\left\{f_{\mu_i}, f_{\alpha_i}\right\}$ are autoregressive neural networks following the approaches in MADE. 

\begin{equation}\label{eq: maf-appendix}
x_i=u_i \exp \alpha_i+\mu_i,
\end{equation}
where 
$\mu_i=f_{\mu_i}\left(\mathbf{x}_{1: i-1}\right), \alpha_i=f_{\alpha_i}\left(\mathbf{x}_{1: i-1}\right)$ and $u_i \sim \mathcal{N}(0,1)$.

\begin{equation}\label{eq: maf-inverse-appendix}
u_i=\left(x_i-\mu_i\right) \exp \left(-\alpha_i\right)
\end{equation}

\begin{equation}\label{eq: maf-jacobian-appendix}
\left|\operatorname{det}J^{-1}\right|=\exp \left(-\sum_i \alpha_i\right)
\end{equation}

Likewise, the inverse autoregressive flow (IAF) \citep{kingma2016improved} uses MADE with Gaussian conditionals and generates data as in Eq. \ref{eq: IAF-appendix}. Its Jacobian determinant has a simple form as in Eq. \ref{eq: IAF-jacobian-appendix}. The main difference between IAF and MAF lies in the history variables. MAF uses previous data variables $\mathbf{x}_{1: i-1}$ to compute $\mu_i$ and $\alpha_i$, whereas IAF uses previous random variables $\mathbf{u}_{1: i-1}$ for the computation. In terms of sampling and density evaluation, IAF can sample in parallel and need to evaluate sequentially, whereas MAF has to sample sequentially and can evaluate in parallel. Since we care more about the sampling efficiency in variational inference, we choose IAF in the paper.

\begin{equation}\label{eq: IAF-appendix}
x_i=u_i \exp \alpha_i+\mu_i,
\end{equation}
where $\mu_i=f_{\mu_i}\left(\mathbf{u}_{1: i-1}\right)$ and $\alpha_i=f_{\alpha_i}\left(\mathbf{u}_{1: i-1}\right)$.

\begin{equation}\label{eq: IAF-jacobian-appendix}
\left|\operatorname{det}J\right|=\exp \left(\sum_i \alpha_i\right)
\end{equation}

\textbf{Affine Coupling} The affine coupling layer, proposed in NICE \citep{dinh2015nice} and later generalized in RealNVP \citep{dinh2017density} takes the following form.

\begin{equation}\label{eq: affine-coupling-appendix}
\begin{cases}y_{1: d} & =x_{1: d} \\ y_{d+1: D} & =x_{d+1: D} \odot \exp \left(s\left(x_{1: d}\right)\right)+t\left(x_{1: d}\right)\end{cases}
\end{equation}
where $s: R^d \mapsto R^{D-d}$ and $t: R^d \mapsto R^{D-d}$ are scale and translation transformation function respectively, and $\odot$ is the element-wise product.

Its Jacobian determinant can be efficiently computed as $\operatorname{det} J = \exp \left[\sum_j s\left(x_{1: d}\right)_j\right]$. Since the computation does not involve the Jacobian of $s$ or $t$, we can make these two functions arbitrarily complex and use neural networks to model them. The coupling layers are usually composed of permutation layers to ensure every component gets modified, and since the Jacobian determinant of permutation is 1, the Jacobian determinant remains tractable.

\textbf{Spline Coupling} Neural spline flows \citep{durkan2019neural, dolatabadi2020invertible} use monotonic rational-quadratic splines or monotonic rational-linear splines as the coupling transformation to achieve more flexibility and yet remain differentiable and invertible. The monotonic rational-quadratic spline uses $K+1$ monotonically increasing knots $\left\{\left(x^{(k)}, y^{(k)}\right)\right\}_{k=0}^K$ to set up $K$ bins, each of which is defined as a rational-quadratic function\footnote{A rational-quadratic function is defined as the quotient of two quadratic polynomial functions.} that is monotonically increasing. It maps $[-B, B]$ to $[-B, B]$ and defines the transformation outside the range to be identity transformation. Let $s_k = \left(y^{k+1}-y^k\right) /\left(x^{k+1}-x^k\right)$ and $\xi(x)=\left(x-x^k\right) /\left(x^{k+1}-x^k\right)$, the rational-quadratic function in the $k$th bin takes the form of Eq. \ref{eq: rational-quadratic-appendix} and the Jacobian determinant of the rational-quadratic neural spline flows (RQNSF) can be written as in Eq. \ref{eq: rqs-jacobian-appendix}.
\begin{equation}\label{eq: rational-quadratic-appendix}
    \resizebox{\columnwidth}{!}{$\frac{\alpha^{(k)}(\xi)}{\beta^{(k)}(\xi)} = y^{(k)}+\frac{\left(y^{(k+1)}-y^{(k)}\right)\left[s^{(k)} \xi^2+\delta^{(k)} \xi(1-\xi)\right]}{s^{(k)}+\left[\delta^{(k+1)}+\delta^{(k)}-2 s^{(k)}\right] \xi(1-\xi)}$}
\end{equation}
\begin{equation}\label{eq: rqs-jacobian-appendix}
\begin{aligned}
& \operatorname{det} J = \prod_{k} \frac{\mathrm{d}}{\mathrm{d} x}\left[\frac{\alpha^{(k)}(\xi)}{\beta^{(k)}(\xi)}\right] \\
= & \resizebox{\columnwidth}{!}{$\prod_{k} \frac{\left(s^{(k)}\right)^2\left[\delta^{(k+1)} \xi^2+2 s^{(k)} \xi(1-\xi)+\delta^{(k)}(1-\xi)^2\right]}{\left[s^{(k)}+\left[\delta^{(k+1)}+\delta^{(k)}-2 s^{(k)}\right] \xi(1-\xi)\right]^2}$}
\end{aligned}
\end{equation}

The rational-linear neural spline flows (RLNSF) work similarly, except with monotonically increasing linear rational functions in each bin. Neural splines combine the best of autoregressive flows and coupling layers (such as NICE and RealNVP) in that it has both an analytic single-pass inverse and sufficient flexibility, as demonstrated in \citet{durkan2019neural}.

\section{Example Analysis}
In this section, we analyze several instances from CNN/Daily Mail and XSum, showcasing diverse outcomes generated by different summarization models.\footnote{It is worth mentioning that a few of the grammatical errors in the summaries can be attributed to the source text itself.}

\begin{table*}[t]
    \centering
	\noindent\fbox{
		\parbox{0.95\linewidth}{
			\textbf{Original Text (truncated):} \textcolor{blue}{It looks like an ordinary forest, with moss climbing up the walls and brown leaves covering the floor. But if you look closely, you will see that this picture is not all it seems. For the peaceful scene actually features a carefully painted female model.} \textcolor{orange}{The amazing illusion is the work of German body-painting artist Joerg Duesterwald, who spent hours painting his model} \textcolor{purple}{so she would blend in with her surroundings.} The \textcolor{orange}{stunning set of pictures was taken in a forest in Langenfeld, Germany, yesterday.} Mr Duesterwald has been painting for more than 20 years. 
			
            \rule{\linewidth}{0.5mm}
            
			\textbf{Gold Summary}: \textcolor{orange}{The illusion is the work of German body-painting artist Joerg Duesterwald, who spent hours painting his model. Stunning set of pictures was taken in front of a rockface in a forest in Langenfeld, Germany, yesterday.}
			
			\rule{\linewidth}{0.1mm}
			
			\textbf{BART}: \textcolor{orange}{Stunning set of images was taken in a forest near Langenfeld, Germany, yesterday by body-painting artist Joerg Duesterwald.} \textcolor{blue}{It looks like an ordinary forest, with moss climbing up the walls and brown leaves covering the floor. But, if you look closely, you will see that this picture is not all it seems. For the peaceful scene actually features a carefully painted female model.}
			
			\rule{\linewidth}{0.1mm}
			
			\textbf{VEDSUM}: \textcolor{orange}{The stunning set of pictures was taken in a forest in Langenfeld, Germany, yesterday.} \textcolor{blue}{It looks like an ordinary forest, with moss climbing up the walls and brown leaves covering the floor. But, if you look closely, you will see that this picture is not all it seems. For the peaceful scene actually features a carefully painted female model.}
			
			\rule{\linewidth}{0.1mm}
			
			\textbf{FlowSUM}: \textcolor{orange}{Amazing illusion is the work of German body-painting artist Joerg Duesterwald. He spent hours painting his model} \textcolor{purple}{so she would blend in with surroundings.} \textcolor{orange}{Stunning set of pictures was taken in a forest in Langenfeld, Germany, yesterday.}
		}
	}
    \caption{\label{tab:cnndm_example_6766}Example 6766 in the CNN/Daily Mail test set: FlowSUM $>$ BART $>$ VEDSUM.}    
\end{table*}

\begin{table*}[t]
    \centering
	\noindent\fbox{
		\parbox{0.95\linewidth}{
			\textbf{Original Text (truncated):} \textcolor{orange}{UFC light heavyweight champion Jon Jones ran from a crash that hospitalised a pregnant woman} - \textcolor{red}{but quickly came back to grab 'a large handful of cash' from the car}, \textcolor{orange}{witnesses told police. According to police, the accident occurred in southeastern Albuquerque just before noon on Sunday local time when the driver of a rented SUV jumped a red light. The driver, whom an off-duty officer identified as Jones, ran from the scene but then returned for the cash before fleeing again, police said.} 'Witnesses stated he shoved the cash into his pants and ran north jumping the fence,' the report said. \textcolor{purple}{Officers found a pipe with marijuana in the vehicle as well as MMA and rental car documents in Jones' name, according to the police report.} Police were searching for UFC  champion Jon Jones in connection with a hit-and-run accident. \textcolor{blue}{Albuquerque police were seeking an arrest warrant for Jones on Monday. They said he would likely face a felony charge of leaving the scene of an accident since the woman broke her arm in the crash.} Police said in a news release they'd been unable to reach Jones or his lawyer. \textcolor{blue}{However, Jones handed himself in later the same day, with TMZ reporting he was being held at Bernalillo County Metropolitan Detention Center.} According to the warrant, the pregnant woman told police she was driving when she was hit by a silver Buick SUV. $\cdots$ Although \textcolor{orange}{he is widely considered the world's best pound-for-pound mixed martial artist}, Jones has endured legal problems and questionable behaviour as champion. 
			
            \rule{\linewidth}{0.5mm}
            
			\textbf{Gold Summary}: \textcolor{orange}{UFC light heavyweight champion Jon Jones ran from a crash that hospitalised a pregnant woman, witnesses told police. According to police, the accident occurred in Albuquerque just before noon on Sunday when the driver of a rented SUV jumped a red light. The driver, whom an off-duty officer identified as Jones, ran from the scene but then returned for the cash before fleeing again, police said. Jones is widely considered the best pound-for-pound mixed martial artist.}
			
			\rule{\linewidth}{0.1mm}
			
			\textbf{BART}: \textcolor{blue}{Albuquerque police were seeking an arrest warrant for Jones on Monday. They said he would likely face a felony charge of leaving the scene of an accident since the woman broke her arm in the crash. However, Jones handed himself in later the same day, withTMZ reporting he was being held at Bernalillo County Metropolitan Detention Center.}
			
			\rule{\linewidth}{0.1mm}
			
			\textbf{VEDSUM}: \textcolor{orange}{UFC light heavyweight champion Jon Jones ran from a crash that hospitalised a pregnant woman.} \textcolor{red}{Witnesses said he returned for 'a large handful of cash' from the car.} \textcolor{blue}{Albuquerque police were seeking an arrest warrant for Jones on Monday. They said he would likely face a felony charge of leaving the scene of an accident since the woman broke her arm in the crash. Jones handed himself in later the same day.}
			
			\rule{\linewidth}{0.1mm}
			
			\textbf{FlowSUM}: \textcolor{orange}{UFC light heavyweight champion Jon Jones ran from a crash that hospitalised a pregnant woman.} \textcolor{red}{Witnesses said he came back to grab 'a large handful of cash' from the car, witnesses told police.} \textcolor{orange}{The driver, whom an off-duty officer identified as Jones, ran from the scene but then returned for the cash before fleeing again, police said.} \textcolor{purple}{Officers found a pipe with marijuana in the vehicle as well as MMA and rental car documents in Jones' name, according to the police report.}
		}
	}
    \caption{\label{tab:cnndm_example_4627}Example 4627 in the CNN/Daily Mail test set: FlowSUM $>$ VEDSUM $>$ BART.}    
\end{table*}

\begin{table*}[t]
    \centering
	\noindent\fbox{
		\parbox{0.95\linewidth}{
			\textbf{Original Text (truncated):} 
   ... \textcolor{orange}{An Icelandic duo has created a snack that is made using cricket flour. Called the Jungle Bar it also contains dates, sesame seeds and chocolate. Cricket flour is said to be a good source of protein and other nutrients. The duo hopes it will encourage people in the West to eat more insects.} \textcolor{red}{The Jungle Bar is being developed by Icelandic duo Búi Bjarmar Aðalsteinsson and Stefán Atli Thoroddsen through their company Crowbar Protein.} \textcolor{blue}{On Kickstarter they are seeking £10,000 (\$15,000) for the insect-powered protein bar.} \textcolor{purple}{They previously rose to fame with their Fly Factory, a micro-factory that used larvae to create foods including chocolate cake and pâté.} Ingredients. Dates, sesame, sunflower and pumpkin seeds, chocolate and cricket flour. Nutrition information. The bar is 50 grams (1.7 ounces), 200 calories, contains 8 grams of high quality protein (16 per cent of the bar) and has a shelf life of 1 year. The duo say that insects are a largely untapped source of nutrients in the Western world, and they hope their product could spark a change in diet habits. \textcolor{red}{To make the bar, the team has cricket flour sent in, from farm-raised crickets that have been ground down.} It is then mixed with the other ingredients to make the unusual bar. 
			
            \rule{\linewidth}{0.5mm}
            
			\textbf{Gold Summary}: \textcolor{orange}{Icelandic duo has created a snack that is made using cricket flour. Called the Jungle Bar it also contains dates, sesame seeds and chocolate. Cricket flour is said to be a good source of protein and other nutrients. The duo hopes it will encourage people in the West to eat more insects.}
			
			\rule{\linewidth}{0.1mm}
			
			\textbf{BART}: \textcolor{orange}{An Icelandic duo has created a snack that is made using cricket flour. Called the Jungle Bar it also contains dates, sesame seeds and chocolate. Cricket flour is said to be a good source of protein and other nutrients. The duo hopes it will encourage people in the West to eat more insects.} \textcolor{blue}{On Kickstarter they are seeking £10,000 (\$15,000) for the insect-powered bar.}
			
			\rule{\linewidth}{0.1mm}
			
			\textbf{VEDSUM}: \textcolor{red}{The Jungle Bar is being developed by Icelandic duo Búi Bjarmar Aðalsteinsson and Stefán Atli Thoroddsen through their company Crowbar Protein.} \textcolor{orange}{Cricket flour is said to be a good source of protein and other nutrients. The duo hopes it will encourage people in the West to eat more insects.}  \textcolor{red}{To make the bar, the team has cricket flour sent in, from farm-raised crickets that have been ground down.}
			
			\rule{\linewidth}{0.1mm}
			
			\textbf{FlowSUM}: \textcolor{orange}{An Icelandic duo has created a snack that is made using cricket flour. Called the Jungle Bar it also contains dates, sesame seeds and chocolate.} \textcolor{red}{The Jungle Bar is being developed by Búi Bjarmar Aðalsteinsson and Stefán Atli Thoroddsen through their company Crowbar Protein.} \textcolor{blue}{On Kickstarter they are seeking £10,000 (\$15,000) for the insect-powered protein bar.} \textcolor{purple}{They previously rose to fame with their Fly Factory, a micro-factory that used larvae to create foods including chocolate cake and pâté.}
		}
	}
    \caption{\label{tab:cnndm_example_4120}Example 4120 in the CNN/Daily Mail test set: BART $>$ VEDSUM $>$ FlowSUM.}
\end{table*}

\begin{table*}[t]
    \centering
	\noindent\fbox{
		\parbox{0.95\linewidth}{
			\textbf{Original Text (truncated):} The Office for National Statistics said industrial output fell 0.7\% compared with January, when it dropped 0.3\%. Unexpectedly warm weather drove the change, because it led to a fall in electricity and gas demand, the ONS said. Construction output fell by 1.7\% in February, down from a revised January reading of zero growth. The construction figure, the biggest drop in nearly a year, was mainly the result of a 2.6\% fall in the housebuilding sector. Meanwhile, the UK's deficit in goods and services widened to Â£3.7bn in February, from a revised figure of Â£3bn in January. According to the ONS, the deficit was fuelled by what it called "erratic items", such as imports of gold and aircraft. "The overall trade deficit worsened, but excluding erratic items, the picture improved, as imports fell more than exports," said ONS senior statistician Kate Davies. Howard Archer, chief UK and European economist at IHS Markit, called the figures "a disappointing package of data for the UK economy which fuels suspicion that GDP growth slowed markedly, largely due to consumers becoming more cautious". He added: "We suspect UK GDP growth in the first quarter of 2017 slowed to 0.4\% quarter-on-quarter from 0.7\% quarter-on-quarter in the fourth quarter of 2016 - this would be the weakest growth rate since the first quarter of 2016."
			
            \rule{\linewidth}{0.5mm}
            
			\textbf{Gold Summary}: Activity in the UK's industrial and construction sectors shrank in February, new figures show.
			
			\rule{\linewidth}{0.1mm}
			
			\textbf{BART}: UK industrial output fell for the second month in a row in February, official figures have shown.
			
			\rule{\linewidth}{0.1mm}
			
			\textbf{VEDSUM}: Industrial output in the UK fell for the second month in a row in February, official figures have shown.
			
			\rule{\linewidth}{0.1mm}
			
			\textbf{FlowSUM}: Activity in the UK's industrial and construction sectors shrank in February, according to official figures.
		}
	}
    \caption{\label{tab:xsum_example_2924}Example 2924 in the XSum test set: FlowSUM $>$ BART $>$ VEDSUM.}
\end{table*}

\begin{table*}[t]
    \centering
	\noindent\fbox{
		\parbox{0.95\linewidth}{
			\textbf{Original Text (truncated):} In December, the government announced finalised plans for a cull, initially in pilot areas, as a way to curb the spread of tuberculosis in cattle. In applying for judicial review, the Badger Trust says culling will not stop TB and may in fact help spread it. Other campaign groups are considering action under the Bern Convention, which protects European wildlife. The government's plans are likely to result in farmers funding contractors to shoot badgers in a number of areas of England, with two initial pilots in west Gloucestershire and west Somerset taking place later this year. "We have identified some serious flaws in the way by which the Secretary of State [Caroline Spelman] reached her decision to cull badgers," said Gwendolen Morgan of Bindmans solicitors, lawyer for the Badger Trust. "Given that Defra's proposals come at an enormous cost to farmers, and threaten to prompt rather than prevent the spread of disease, we hope that this ill-conceived decision will be struck down by the court." She pointed to government projections that culling would reduce TB incidence by 12-16\% over nine years.
			
            \rule{\linewidth}{0.5mm}
            
			\textbf{Gold Summary}: The Badger Trust has launched a new legal challenge to the government's plans to cull badgers in England.
   
			\rule{\linewidth}{0.1mm}
			
			\textbf{BART}: The Badger Trust has launched a legal challenge to the government's plans to cull badgers in England.
			
			\rule{\linewidth}{0.1mm}
			
			\textbf{VEDSUM}: The Badger Trust is taking legal action against the Department for Environment, Food and Rural Affairs (Defra) over plans to cull badgers in England.
			
			\rule{\linewidth}{0.1mm}
			
			\textbf{FlowSUM}: The Badger Trust has launched a legal challenge to the UK government's plans to cull badgers in England and Wales.
		}
	}
    \caption{\label{tab:xsum_example_5737}Example 5737 in the XSum test set: BART $>$ FlowSUM $>$ VEDSUM.}
\end{table*}

\begin{table*}[t]
    \centering
	\noindent\fbox{
		\parbox{0.95\linewidth}{
			\textbf{Original Text (truncated):} The response from many in that time has been: "Let's get on with it." That view was shared by the First Minister Carwyn Jones until recently when he altered his opinion and said that we should only start the official Brexit negotiations in the early part of next year. My sense is that the public will be flexible on the timing up to a point, as long as they are given a clear sense of direction. The majority of the political establishment have had to come to terms with the fact that most people ignored their advice to remain. So much for being in touch with the electorate. In conversations with politicians on the remain side since, I have come across a mix of bewilderment, frustration and sadness. And while people like me spend a lot of time talking and writing about a Welsh political dynamic, on this subject at least, Wales was a carbon copy of England. In stark contrast, those that supported leaving feel vindicated by their campaign, and now believe they are the ones in touch with vast swathes of the population. The referendum result was a devastating indictment of the effectiveness of the billions of pounds of EU funds spent trying to regenerate economically deprived communities. The brutal reality is that those who were most likely to vote to leave lived in communities where most EU money had been spent. It is an extraordinary paradox that raised eyebrows far further afield than Wales. 
			
            \rule{\linewidth}{0.5mm}
            
			\textbf{Gold Summary}: It has been a month since Wales voted to leave the European Union. 
			
			\rule{\linewidth}{0.1mm}
			
			\textbf{BART}: It has been more than a year since the UK voted to leave the European Union.

			\rule{\linewidth}{0.1mm}
			
			\textbf{VEDSUM}: It has been a year since the EU referendum result, and in that time I have spent a great deal of time talking to politicians on both sides of the political spectrum about what they think about Brexit.

			\rule{\linewidth}{0.1mm}
			
			\textbf{FlowSUM}: Since the referendum result on 23 June, I have spent a lot of time talking about the implications for Wales and the Welsh political establishment. 
		}
	}
    \caption{\label{tab:xsum_example_9512}Example 9512 in the XSum test set: BART $>$ VEDSUM $>$ FlowSUM.}
\end{table*}

\end{document}